\PassOptionsToPackage{unicode}{hyperref}
\PassOptionsToPackage{hyphens}{url}
\documentclass[
]{article}
\usepackage{xcolor}
\usepackage{amsmath,amssymb}
\usepackage{iftex}
\ifPDFTeX
  \usepackage[T1]{fontenc}
  \usepackage[utf8]{inputenc}
  \usepackage{textcomp} 
\else 
  \usepackage{unicode-math} 
  \defaultfontfeatures{Scale=MatchLowercase}
  \defaultfontfeatures[\rmfamily]{Ligatures=TeX,Scale=1}
\fi
\usepackage{lmodern}
\ifPDFTeX\else
\fi
\IfFileExists{upquote.sty}{\usepackage{upquote}}{}
\IfFileExists{microtype.sty}{
  \usepackage[]{microtype}
  \UseMicrotypeSet[protrusion]{basicmath} 
}{}
\makeatletter
\@ifundefined{KOMAClassName}{
  \IfFileExists{parskip.sty}{%
    \usepackage{parskip}
  }{
    \setlength{\parindent}{0pt}
    \setlength{\parskip}{6pt plus 2pt minus 1pt}}
}{
  \KOMAoptions{parskip=half}}
\makeatother
\usepackage{longtable,booktabs,array}
\usepackage{calc} 
\usepackage[
  left=1in,
  right=1in,
  top=1in,
  bottom=1in,
  paper=letterpaper
]{geometry}

\usepackage{setspace}
\usepackage{subcaption}
\usepackage{etoolbox}
\makeatletter
\patchcmd\longtable{\par}{\if@noskipsec\mbox{}\fi\par}{}{}
\makeatother
\IfFileExists{footnotehyper.sty}{\usepackage{footnotehyper}}{\usepackage{footnote}}
\makesavenoteenv{longtable}
\usepackage{graphicx}
\makeatletter
\newsavebox\pandoc@box
\newcommand*\pandocbounded[1]{
  \sbox\pandoc@box{#1}%
  \Gscale@div\@tempa{\textheight}{\dimexpr\ht\pandoc@box+\dp\pandoc@box\relax}%
  \Gscale@div\@tempb{\linewidth}{\wd\pandoc@box}%
  \ifdim\@tempb\p@<\@tempa\p@\let\@tempa\@tempb\fi
  \ifdim\@tempa\p@<\p@\scalebox{\@tempa}{\usebox\pandoc@box}%
  \else\usebox{\pandoc@box}%
  \fi%
}
\def\fps@figure{htbp}
\makeatother
\usepackage{bookmark}
\IfFileExists{xurl.sty}{\usepackage{xurl}}{} 
\usepackage[numbers,square]{natbib}
\usepackage{tikz}
\usetikzlibrary{arrows.meta, positioning, mindmap}
\usepackage{pgfplots}
\pgfplotsset{compat=1.18}
\usepackage{subcaption}
\usepackage{float}
\usepackage[font=small,labelfont=bf,textfont=normal,justification=centering]{caption}

\tikzset{
  stage/.style   = {draw, rectangle, rounded corners=2pt, align=center,
                    minimum width=7cm, inner sep=6pt, font=\small},
  arrow/.style   = {-{Stealth[length=6pt]}, line width=0.6pt},
  tiny/.style    = {font=\scriptsize, align=left}
}

\hypersetup{
  pdftitle={Advancing Offline Handwritten Text Recognition: A Systematic Review of Data Augmentation and Generation Techniques},
  colorlinks=true,
  citecolor=blue,
  linkcolor=black,
  urlcolor=blue,
  pdfcreator={LaTeX via pandoc}}

\title{Advancing Offline Handwritten Text Recognition: A Systematic Review of Data Augmentation and Generation Techniques}

\author{Yassin Hussein Rassul\textsuperscript{1,*}, 
        Aram M. Ahmed\textsuperscript{2}, 
        Dr. Polla Fattah\textsuperscript{2,5}, 
        Bryar A. Hassan\textsuperscript{2,4}, \\
        Arwaa W. Abdulkareem\textsuperscript{1}, 
        Tarik A. Rashid\textsuperscript{1,2}, 
        Joan Lu\textsuperscript{3}}

\date{}

\begin{document}
\maketitle

\begin{center}
\small
\vspace{0.5em}
\textsuperscript{1}Artificial Intelligence and Innovation Centre, University of Kurdistan Hewler, Erbil, Iraq\\[0.4em]

\textsuperscript{2}Computer Science and Engineering Department, School of Science and Engineering,\\ 
University of Kurdistan Hewler, Erbil, Iraq\\[0.4em]

\textsuperscript{3}School of Computing and Engineering, University of Huddersfield, Huddersfield HD1 3DH, UK\\[0.4em]

\textsuperscript{4}Department of Computer Science, College of Science, Charmo University,\\ 
46023 Chamchamal, Sulaimani, Iraq\\[0.4em]

\textsuperscript{5}Software and Informatics Engineering, College of Engineering, Salahaddin University-Erbil, Iraq\\[0.8em]

\textsuperscript{*}Corresponding author: \href{mailto:yassin.hussein@ukh.edu.krd}{yassin.hussein@ukh.edu.krd}
\vspace{0.5em}
\end{center}

\vspace{1em}

\noindent\textbf{Abstract}

\vspace{0.5em}

Offline Handwritten Text Recognition (HTR) systems play a crucial role
in applications such as historical document digitization, automatic form
processing, and biometric authentication. However, their performance is
often hindered by the limited availability of annotated training data,
particularly for low-resource languages and complex scripts. This paper
presents a comprehensive survey of offline handwritten data augmentation
and generation techniques designed to improve the accuracy and
robustness of HTR systems. We systematically examine traditional
augmentation methods alongside recent advances in deep learning,
including Generative Adversarial Networks (GANs), diffusion models, and
transformer-based approaches. Furthermore, we explore the challenges
associated with generating diverse and realistic handwriting samples,
particularly in preserving script authenticity and addressing data
scarcity. This survey follows the PRISMA methodology, ensuring a
structured and rigorous selection process. Our analysis began with 1,302
primary studies, which were filtered down to 848 after removing
duplicates, drawing from key academic sources such as IEEE Digital
Library, Springer Link, Science Direct, and ACM Digital Library. By
evaluating existing datasets, assessment metrics, and state-of-the-art
methodologies, this survey identifies key research gaps and proposes
future directions to advance the field of handwritten text generation
across diverse linguistic and stylistic landscapes.

\vspace{0.5em}

\noindent\textbf{Keywords:} Data augmentation $\bullet$ Handwriting synthesis $\bullet$ Handwritten text recognition $\bullet$ Generative adversarial networks $\bullet$ Systematic literature review

\section{Introduction}\label{introduction}

Handwritten Text Recognition (HTR) has become an essential tool for
applications such as historical document digitization, automatic form
processing, and biometric authentication. Despite significant
advancements, these systems often face challenges due to the limited
availability of annotated training data, particularly for low-resource
languages and complex scripts. High variability in handwriting styles,
the need for script preservation, and the computational demands of deep
learning models further complicate the development of robust HTR systems
\cite{boteanu2023}.

To address these challenges, handwritten data augmentation and
generation techniques have emerged as crucial solutions. Traditional
approaches, such as geometric transformations and noise injections, have
been widely used to enhance dataset diversity \cite{madaan2022}. However, recent
advancements in deep learning, particularly Generative Adversarial
Networks (GANs), diffusion models, and transformer-based architectures,
have introduced more sophisticated methods for synthesizing realistic
handwritten text. These techniques not only expand datasets but also
improve recognition performance across diverse writing styles and
languages \cite{aksan2018}.

This paper presents a comprehensive survey of offline handwritten data
augmentation and generation methods, systematically evaluating their
effectiveness in enhancing HTR systems. Using the PRISMA methodology
\cite{moher2009} \cite{kitchenham2009}, we analyzed 1,302 primary studies, filtering them down
to 848 after removing duplicates, sourced from IEEE Digital Library,
Arxiv, Springer Link, Science Direct, and ACM Digital Library. For
managing and analyzing the papers, a combination of Zotero and a custom
spreadsheet was used to keep track of the process. This review examines
existing datasets, evaluation metrics, and state-of-the-art
methodologies, highlighting key challenges and identifying future
research directions to advance the field.

The structure of this paper is as follows: Section 2 describes the
research methodology, including the PRISMA approach and meta-results.
Section 3 presents the key findings, covering data augmentation methods,
datasets, and evaluation metrics. Section 4 discusses challenges and
proposed solutions in handwritten text generation. Section 5 provides a
detailed discussion, addressing each research question and exploring the
evolution and effectiveness of techniques. Finally, Section 6 concludes
the study, summarizing the main contributions and suggesting future
research directions to advance offline handwritten text recognition.

\section{Review Methodology}\label{review-methodology}

This section outlines the methodology employed in conducting
the survey, which includes the PRISMA approach, and the meta results
obtained.

\subsection{PRISMA Approach}\label{prisma-approach}

This survey followed the PRISMA (Preferred Reporting Items for
Systematic Reviews and Meta-Analyses) methodology \cite{moher2009}, which is a
set of guidelines designed to enhance transparency, consistency, and
quality of reporting in systematic reviews and meta-analyses. Therefore,
the research was conducted in four distinct phases, namely
Identification, Screening, in-depth review, and Quality Criteria (QC)
filtering. Figure 1 shows a sequential view of this process.

\begin{figure}[H]
\centering
\begin{tikzpicture}[node distance=10mm]
  \node[font=\bfseries, align=center] (title) {PRISMA Overview – systematic review process};

  \node[stage, below=5mm of title] (sources) {Databases Searched\\ACM • IEEE • ScienceDirect\\SpringerLink • arXiv};

  \node[stage, below=10mm of sources] (id) {Records identified: \textbf{1\,302}};
  \node[tiny, right=6mm of id] (q1) {Search string: \emph{"handwritten text generation"} etc.};

  \node[stage, below=10mm of id] (dedup) {After duplicates removed: \textbf{848}\\\scriptsize (454 exact duplicates excluded)};

  \node[stage, below=10mm of dedup] (screen) {Title \& abstract screening: \textbf{91} retained\\\scriptsize (757 excluded by criteria EC1-EC7)};
  \node[tiny, right=6mm of screen] (ic) {Key Inclusion (IC1):\\• Offline handwritten text\\• Generative models, 2010\,+\\• English full-text};
  \node[tiny, left=6mm of screen]  (ec) {Key Exclusion (EC1-EC7):\\• Printed-text focus\\• Real-time handwriting\\• Pre-2010, obsolete tech\\• Non-English, duplicates};

  \node[stage, below=10mm of screen] (full) {Full-text review: \textbf{55} retained\\\scriptsize (36 excluded; EC8 = quality < 7/11)};

  \node[stage, below=10mm of full] (qual) {Quality assessment (11-item checklist)\\All \textbf{55} meet $\geq$7 items};

  \node[stage, below=10mm of qual] (final) {\textbf{Final corpus: 55 papers}\\Breakdown – ACM 12 • IEEE 13 • ScienceDirect 5 • Springer 7 • arXiv 18};

  \draw[arrow] (sources) -- (id);
  \draw[arrow] (id)      -- (dedup);
  \draw[arrow] (dedup)   -- (screen);
  \draw[arrow] (screen)  -- (full);
  \draw[arrow] (full)    -- (qual);
  \draw[arrow] (qual)    -- (final);
\end{tikzpicture}
\caption{PRISMA flow diagram showing the systematic review process for offline handwritten text generation studies.}
\end{figure}
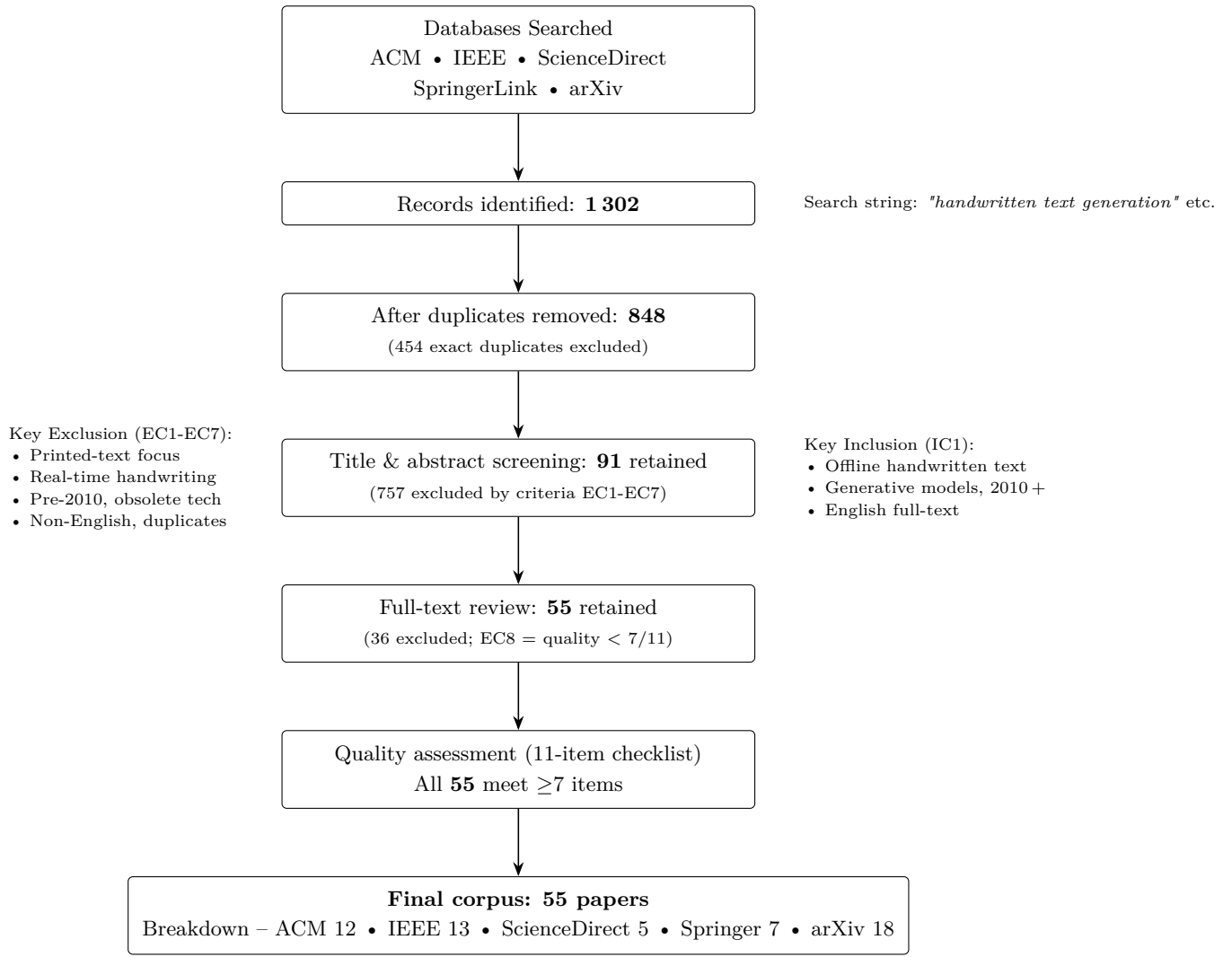

\textbf{Phase 1: Identification}

At this stage the research questions and objectives are first defined.
Next, the Academic Databases are selected. Then a search strategy is
developed to collect the articles.

Initially, the review seeks to address the following key questions:

-- \textbf{RQ1:} How have techniques for augmenting and generating
offline handwritten data evolved, and what are the current leading
methods?

-- \textbf{RQ2:} How do these techniques compare in terms of
effectiveness and efficiency for improving OCR accuracy and overall
recognition performance?

-- \textbf{RQ3:} What are the main challenges in creating realistic and
diverse synthetic handwriting samples, and how can these challenges be
overcome?

-- \textbf{RQ4:} How can these techniques be used to expand and
diversify datasets, especially for languages with limited resources?

-- \textbf{RQ5:} Which datasets are most commonly used and most
effective for augmenting and generating offline handwritten data?

-- \textbf{RQ6:} How can we assess the quality and variety of augmented
and generated handwriting, particularly for languages with limited
existing datasets?

Accordingly, this survey aims to: (1) explore existing literature on data augmentation techniques for offline handwritten text recognition; (2) identify specific data augmentation methods applied to offline handwritten text recognition; (3) focus on studies that target cursive text images, as they are relevant to the objective of handwriting recognition research; and (4) exclude studies related to online and printed text recognition, scene text, digits, signatures, and mathematical expression recognition.

The following academic databases were selected for this study: ACM
Digital Library, IEEE Digital Library, ScienceDirect, arXiv, and
SpringerLink. These databases were chosen due to their comprehensive
coverage of high-quality research publications in the field. Notably,
arXiv was included as it serves as a primary repository for preprint
studies, where many of the most recent advancements in offline
handwritten text recognition are first published, ensuring access to the
latest developments in the field.

As to the search strategy, several filtering criteria, including time
period, subject area, document type, publication status, and language,
were applied directly within the search engines. However, as filtering
options varied across platforms, any missing criteria were manually
applied in the subsequent step. So, a predefined set of keywords were
inserted into the selected databases, with the search restricted to
studies published between 2010 and 2024, covering a 14-year period.
Table 1 presents the selected keywords along with their corresponding
synonyms.

\begin{longtable}{p{0.45\textwidth}p{0.45\textwidth}}
\caption{Keywords and synonyms used to generate the search string}\\
\toprule
\textbf{Keywords} & \textbf{Synonyms} \\
\midrule
\endfirsthead
\toprule
\textbf{Keywords} & \textbf{Synonyms} \\
\midrule
\endhead
\bottomrule
\endlastfoot
Handwritten text generation & Handwriting synthesis, handwriting generation \\
text-to-handwritten image synthesis & offline handwriting generation, offline handwriting synthesis \\
historical document recreation & classic document imitation, calligraphy emulation, ancient script simulation \\
\end{longtable}

Based on these keywords, the following search string was utilized to
retrieve relevant studies: \emph{("handwritten text generation" OR
"handwriting synthesis" OR "handwriting generation" OR
"text-to-handwritten image synthesis" OR "offline handwriting
generation" OR "offline handwriting synthesis" OR "historical document
recreation" OR "handwriting augmentation").}

\textbf{Phase 2: Screening:}

At this stage, a preliminary screening was conducted by briefly
reviewing the titles and abstracts of the retrieved studies. Two
independent reviewers assessed each study based on predefined inclusion
and exclusion criteria. If at least one reviewer deemed a study
relevant, it proceeded to the next stage. Therefore, the following
exclusion criteria were used to filter out articles that
don't fit the scope of this review:

\begin{quote}
\textbf{EC1}: Articles that focus on printed text generation are
excluded.

\textbf{EC2}: Articles related to real-time or online handwriting input
are excluded.

\textbf{EC3}: Articles published before 2010 are excluded.

\textbf{EC4}: Articles that do not involve generative models or
techniques for creating new handwritten text are excluded.

\textbf{EC5}: Articles that rely on outdated or obsolete technologies
not relevant to current advancements are excluded.

\textbf{EC6}: Articles published in languages other than English are
excluded.

\textbf{EC7}: Duplicate work is excluded.

\textbf{EC8}: Works that do not meet at least seven out of our defined
quality standards are excluded.
\end{quote}

Then, the revivers applied the below Inclusion Criteria (IC):

\begin{quote}
\textbf{IC1:} Studies on offline handwritten text generation, published
since 2010, that use generative models and modern technologies, are
written in English, are not duplicates, and involve data augmentation
\end{quote}

\textbf{Phase 3: In-depth Review}

At this stage, a comprehensive full-text review of the remaining studies
was conducted, applying the same inclusion and exclusion criteria to
ensure alignment with the research objectives. Notably, the eight
standard threshold was established to ensure that only studies with
strong technical and descriptive quality are considered, as highlighted
by \cite{kitchenham2010}. finally, it is worth mentioning that the searches were done
using the advanced search features of each platform, looking at both
metadata and the full text of papers.

\textbf{Phase 4: Quality Criteria Filtering}

At this stage, the studies were scored by the reviewers using the
following Quality Criteria (QC):

\begin{quote}
\textbf{QC1}: Are the research objectives clearly explained?

\textbf{QC2}: Is the methodology well described?

\textbf{QC3}: Is the data augmentation or generation method clearly
explained?

\textbf{QC4}: Are the performance metrics for handwriting recognition
well defined?

\textbf{QC5}: Does the study compare its approach to existing methods?

\textbf{QC6}: Are the results properly analyzed and discussed?

\textbf{QC7}: Does the article mention any limitations or challenges of
the method?

\textbf{QC8}: Does the study consider different languages or scripts?

\textbf{QC9}: Is the code or pseudocode provided for others to
reproduce?

\textbf{QC10}: Is the dataset clearly described?

\textbf{QC11}: Is the source code made available to the public?
\end{quote}

Each question was assessed using a binary scoring system, where
responses were assigned 0 points for "No" and 1 point for "Yes." The
final score for each study was then evaluated using EC8, as detailed in
the Research Questions and Strategy section. Notably, no studies were
excluded, as all met at least seven of the predefined Quality Criteria.

\subsection{Meta Data Results}\label{meta-data-results}

This section presents the quantitative outcomes of our systematic review process, detailing the study selection and filtering procedures across all four PRISMA phases. The results demonstrate the effectiveness of our search strategy and quality assessment criteria in identifying relevant literature in the field of offline handwritten text generation and augmentation.

\subsubsection{Identification and
Screening}\label{identification-and-screening}

A total of 1,302 primary studies were initially gathered from five
sources: IEEE Digital Library (31), Arxiv (290), Springer Link (231),
Science Direct (262), and ACM Digital Library (34). After removing
duplicates, 848 unique records remained. At this early stage, only the
default database filters were applied (see previous section for
details). The distribution of studies across these databases is
illustrated in Figure 2.

A preliminary screening of titles and abstracts followed, guided by the
eight exclusion criteria (EC) described in previous section. This step
aimed to eliminate works clearly irrelevant to offline handwritten text
generation or lacking data augmentation components. From the 848 unique
papers, 91 were retained as potentially relevant for further review.

\subsubsection{Full-Text Review and Quality
Assessment}\label{full-text-review-and-quality-assessment}

Next, the 91 remaining papers underwent full-text review to remove false
positives. Studies focusing solely on handwriting recognition or
scene-text recognition—without any generative or augmentation
component—were excluded. This phase narrowed the corpus to 55 papers,
each of which was then assessed against predefined quality criteria
(QC).

All 55 remaining papers satisfied EC8, meaning they met at least 7
of the designated QCs, thereby demonstrating acceptable methodological
rigor.

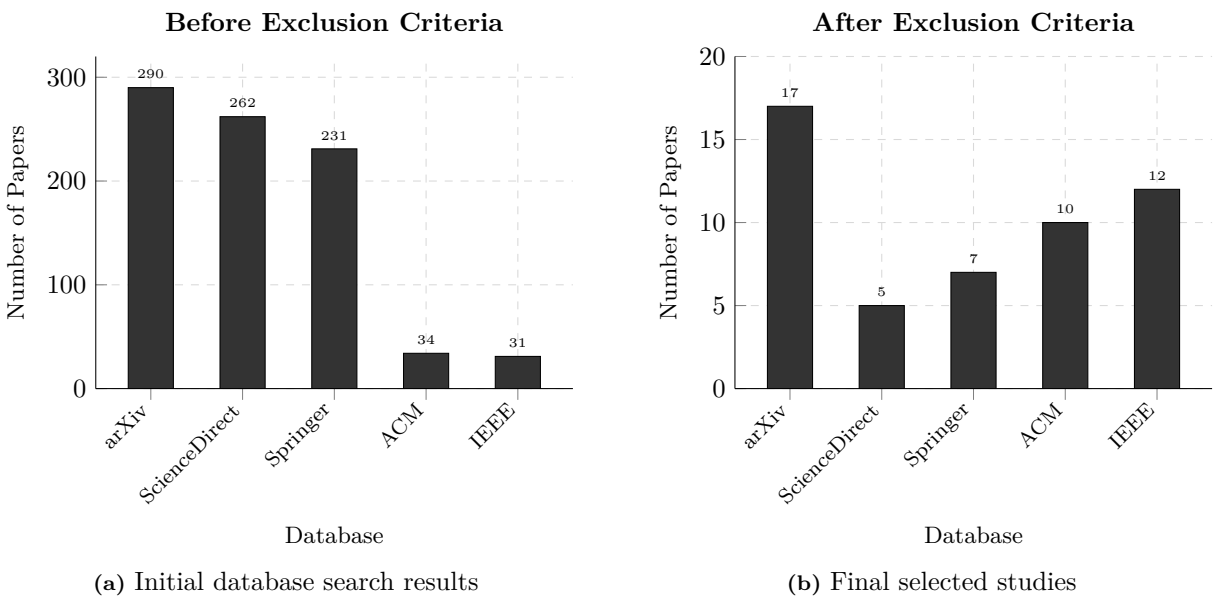
\begin{figure}[htbp]
\centering
\begin{subfigure}[b]{0.48\textwidth}
  \centering
  \begin{tikzpicture}
    \begin{axis}[
        ybar,
        width=\textwidth,
        height=6cm,
        bar width=0.6cm,
        ymin=0,
        ymax=320,
        ylabel={Number of Papers},
        xlabel={Database},
        xtick=data,
        symbolic x coords={arXiv, ScienceDirect, Springer, ACM, IEEE},
        xticklabel style={rotate=45, anchor=east, font=\footnotesize},
        nodes near coords,
        every node near coord/.append style={font=\tiny},
        grid=major,
        grid style={dashed, gray!30},
        axis lines*=left,
        ylabel style={font=\small},
        xlabel style={font=\small},
        enlarge x limits=0.15,
        title={Before Exclusion Criteria},
        title style={font=\normalsize\bfseries},
      ]
      \addplot[ybar, fill=black!80] coordinates {
        (arXiv, 290)
        (ScienceDirect, 262)
        (Springer, 231)
        (ACM, 34)
        (IEEE, 31)
      };
    \end{axis}
  \end{tikzpicture}
  \caption{Initial database search results}
  \label{fig:databases_before}
\end{subfigure}
\hfill
\begin{subfigure}[b]{0.48\textwidth}
  \centering
  \begin{tikzpicture}
    \begin{axis}[
        ybar,
        width=\textwidth,
        height=6cm,
        bar width=0.6cm,
        ymin=0,
        ymax=20,
        ylabel={Number of Papers},
        xlabel={Database},
        xtick=data,
        symbolic x coords={arXiv, ScienceDirect, Springer, ACM, IEEE},
        xticklabel style={rotate=45, anchor=east, font=\footnotesize},
        nodes near coords,
        every node near coord/.append style={font=\tiny},
        grid=major,
        grid style={dashed, gray!30},
        axis lines*=left,
        ylabel style={font=\small},
        xlabel style={font=\small},
        enlarge x limits=0.15,
        title={After Exclusion Criteria},
        title style={font=\normalsize\bfseries},
      ]
      \addplot[ybar, fill=black!80] coordinates {
        (arXiv, 17)
        (ScienceDirect, 5)
        (Springer, 7)
        (ACM, 10)
        (IEEE, 12)
      };
    \end{axis}
  \end{tikzpicture}
  \caption{Final selected studies}
  \label{fig:databases_after}
\end{subfigure}
\caption{The distribution of studies across databases before and after applying exclusion criteria (EC). The initial search yielded 848 unique papers (a), which were reduced to 51 relevant studies after applying the exclusion criteria (b).}
\label{fig:database_distribution}
\end{figure}

\section{Collected Studies and Key
Findings}\label{collected-studies-and-key-findings}

In this section, we present an overview of the papers
collected during the survey. The selected studies are classified into
three groups, namely Data augmentation and generation methods, The
available datasets for data augmentation and generation, and finally
Evaluation Metrics and Benchmarks.

\subsection{Data augmentation and generation
methods}\label{data-augmentation-and-generation-methods}

The field of handwritten text generation has seen considerable
advancements, particularly with the evolution of machine learning and
deep learning techniques. This section explores various methodologies
for generating synthetic handwritten text, systematically grouped based
on their underlying attributes and methods. Each approach is discussed
in terms of its impact on the field, addressing its innovations,
challenges, and applications. Figure 3 and 4 show a summary of the
methodological advancements in handwritten generation as well as
existing methods in the literature.

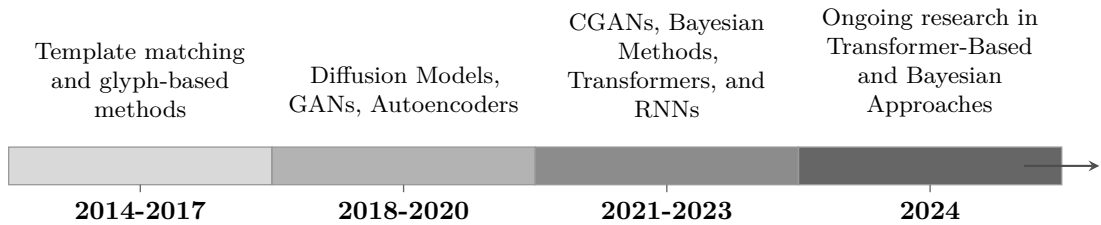
\begin{figure}
\centering
\begin{tikzpicture}[scale=1]
    
    \draw[line width=0.8pt, black!70] (0,0) -- (14,0);
    
    \fill[black!15] (0,-0.25) rectangle (3.5,0.25);
    \fill[black!30] (3.5,-0.25) rectangle (7,0.25);
    \fill[black!45] (7,-0.25) rectangle (10.5,0.25);
    \fill[black!60] (10.5,-0.25) rectangle (14,0.25);
    
    \draw[line width=0.5pt, black!40] (0,-0.25) rectangle (3.5,0.25);
    \draw[line width=0.5pt, black!40] (3.5,-0.25) rectangle (7,0.25);
    \draw[line width=0.5pt, black!40] (7,-0.25) rectangle (10.5,0.25);
    \draw[line width=0.5pt, black!40] (10.5,-0.25) rectangle (14,0.25);
    
    \node[below=10pt, font=\normalsize] at (1.75,0) {\textbf{2014-2017}};
    \node[below=10pt, font=\normalsize] at (5.25,0) {\textbf{2018-2020}};
    \node[below=10pt, font=\normalsize] at (8.75,0) {\textbf{2021-2023}};
    \node[below=10pt, font=\normalsize] at (12.25,0) {\textbf{2024}};
    
    \node[above=15pt, text width=3.2cm, align=center, font=\small] at (1.75,0) {
        Template matching\\
        and glyph-based\\
        methods
    };
    
    \node[above=15pt, text width=3.2cm, align=center, font=\small] at (5.25,0) {
        Diffusion Models,\\
        GANs, Autoencoders
    };
    
    \node[above=15pt, text width=3.4cm, align=center, font=\small] at (8.75,0) {
        CGANs, Bayesian\\
        Methods,\\
        Transformers, and\\
        RNNs
    };
    
    \node[above=15pt, text width=3.2cm, align=center, font=\small] at (12.25,0) {
        Ongoing research in\\
        Transformer-Based\\
        and Bayesian\\
        Approaches
    };
    
    \draw[line width=0.5pt, black!60] (1.75,-0.25) -- (1.75,-0.35);
    \draw[line width=0.5pt, black!60] (5.25,-0.25) -- (5.25,-0.35);
    \draw[line width=0.5pt, black!60] (8.75,-0.25) -- (8.75,-0.35);
    \draw[line width=0.5pt, black!60] (12.25,-0.25) -- (12.25,-0.35);
    
    \draw[line width=0.8pt, ->, >=stealth, black!70] (13.5,0) -- (14.5,0);
    
\end{tikzpicture}
\caption{Timeline of Methodological Advancements in Handwritten Generation}
\end{figure}

\clearpage
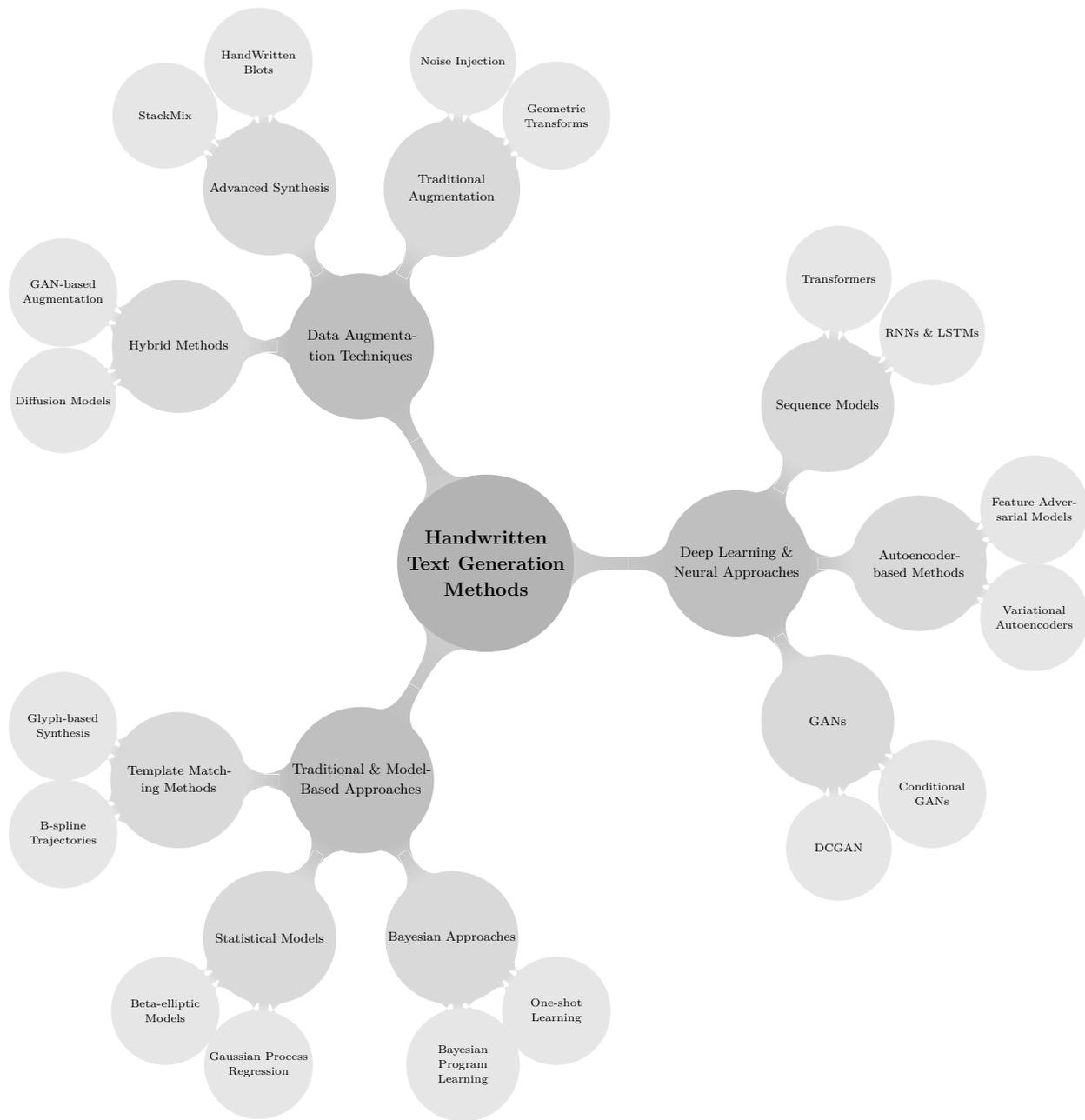
\begin{figure}[H]
\centering
\resizebox{\textwidth}{!}{%
\begin{tikzpicture}[mindmap, grow cyclic, every node/.style=concept, concept color=black!20, 
    level 1/.append style={level distance=5.5cm,sibling angle=120},
    level 2/.append style={level distance=4cm,sibling angle=60},
    level 3/.append style={level distance=2.8cm,sibling angle=50},
    font=\small,
    every concept/.append style={minimum size=0pt, text width=auto}]

\node[concept, concept color=black!30, font=\large\bfseries, text width=3.5cm, align=center] {Handwritten Text Generation Methods}
    child[concept color=black!25] { node[text width=3cm, align=center] {Traditional \& Model-Based Approaches}
        child[concept color=black!15] { node[text width=2.8cm, align=center, font=\footnotesize] {Template Matching Methods} 
            child[concept color=black!10] { node[text width=2.2cm, font=\scriptsize] {Glyph-based Synthesis} }
            child[concept color=black!10] { node[text width=2.2cm, font=\scriptsize] {B-spline Trajectories} }
        }
        child[concept color=black!15] { node[text width=2.8cm, align=center, font=\footnotesize] {Statistical Models}
            child[concept color=black!10] { node[text width=2.2cm, font=\scriptsize] {Beta-elliptic Models} }
            child[concept color=black!10] { node[text width=2.2cm, font=\scriptsize] {Gaussian Process Regression} }
        }
        child[concept color=black!15] { node[text width=2.8cm, align=center, font=\footnotesize] {Bayesian Approaches}
            child[concept color=black!10] { node[text width=2.2cm, font=\scriptsize] {Bayesian Program Learning} }
            child[concept color=black!10] { node[text width=2.2cm, font=\scriptsize] {One-shot Learning} }
        }
    }
    child[concept color=black!25] { node[text width=3cm, align=center] {Deep Learning \& Neural Approaches}
        child[concept color=black!15] { node[text width=2.8cm, align=center, font=\footnotesize] {GANs}
            child[concept color=black!10] { node[text width=2.2cm, font=\scriptsize] {DCGAN} }
            child[concept color=black!10] { node[text width=2.2cm, font=\scriptsize] {Conditional GANs} }
        }
        child[concept color=black!15] { node[text width=2.8cm, align=center, font=\footnotesize] {Autoencoder-based Methods}
            child[concept color=black!10] { node[text width=2.2cm, font=\scriptsize] {Variational Autoencoders} }
            child[concept color=black!10] { node[text width=2.2cm, font=\scriptsize] {Feature Adversarial Models} }
        }
        child[concept color=black!15] { node[text width=2.8cm, align=center, font=\footnotesize] {Sequence Models}
            child[concept color=black!10] { node[text width=2.2cm, font=\scriptsize] {RNNs \& LSTMs} }
            child[concept color=black!10] { node[text width=2.2cm, font=\scriptsize] {Transformers} }
        }
    }
    child[concept color=black!25] { node[text width=3cm, align=center] {Data Augmentation Techniques}
        child[concept color=black!15] { node[text width=2.8cm, align=center, font=\footnotesize] {Traditional Augmentation}
            child[concept color=black!10] { node[text width=2.2cm, font=\scriptsize] {Geometric Transforms} }
            child[concept color=black!10] { node[text width=2.2cm, font=\scriptsize] {Noise Injection} }
        }
        child[concept color=black!15] { node[text width=2.8cm, align=center, font=\footnotesize] {Advanced Synthesis}
            child[concept color=black!10] { node[text width=2.2cm, font=\scriptsize] {HandWritten Blots} }
            child[concept color=black!10] { node[text width=2.2cm, font=\scriptsize] {StackMix} }
        }
        child[concept color=black!15] { node[text width=2.8cm, align=center, font=\footnotesize] {Hybrid Methods}
            child[concept color=black!10] { node[text width=2.2cm, font=\scriptsize] {GAN-based Augmentation} }
            child[concept color=black!10] { node[text width=2.2cm, font=\scriptsize] {Diffusion Models} }
        }
    };

\end{tikzpicture}%
}
\caption{Taxonomy Diagram of Handwritten Text Generation Methods}
\end{figure}
\clearpage

\subsubsection{Traditional and Model-Based
Approaches}\label{traditional-and-model-based-approaches}

Several studies have explored traditional, model-based strategies for
specific tasks, such as Arabic writer identification. For instance, the
paper \cite{abdi2015} implemented a beta-elliptic model to construct synthetic
graphemes for Arabic writer identification/verification. Their approach
utilizes synthetic codebooks as templates for feature extraction, which
enhances robustness and generalization. Despite the model's limited
sensitivity to particular handwriting styles, it achieved a Top1 rate of
$90.02\%$ and an Equal Error Rate (EER) of $2.1\%$, outperforming earlier
methods.

Template matching and glyph-based methods have also been investigated.
Synthesized chinese calligraphy has been proposed in \cite{li2014} by
extracting strokes from a limited set of calligraphic samples and
matching them to user-specified B-spline trajectories. This method
maintains stylistic accuracy through innovations such as a Weighted
F-histogram for topology representation and Adaboost with Support Vector
Regressors (SVRs) for style evaluation, although it faces challenges in
scalability and computational efficiency. Similarly, research in \cite{jiang2017}
introduced DCFont, a deep learning--based system for generating
personalized Chinese font libraries. The integration of adversarial
training and an end-to-end framework in DCFont minimizes human
intervention, yet the method requires extensive training data and
encounters difficulties with cursive characters.

Another notable model-based approach is presented by \cite{dey2016} which
proposed generating synthetic handwriting using n-gram letter glyphs
combined with non-parametric Gaussian Process (GP) regression and
dynamic programming for n-gram parsing. This method enhances realism in
synthetic handwriting, making it useful for training recognition systems
and personalization, though its computational demands restrict its
application to smaller datasets. Additionally, in \cite{balreira2017} demonstrates
a method that matches individual handwriting styles using publicly
available fonts, incorporating preprocessing, Procrustes distance for
shape matching, and distance transform values for thickness matching,
followed by synthesis via glyph concatenation. This technique strikes a
balance between visual similarity and processing efficiency, achieving a
similarity score of 7.1/10, but it relies on manual text segmentation,
which may limit scalability.

Bayesian and one-shot learning methods have also been applied. Research
in \cite{souibgui2021} utilizes Bayesian Program Learning (BPL) to generate
synthetic handwritten data from low-resource alphabets. BPL's
hierarchical sampling from primitives—augmented by transformations
such as rotation and resizing—reduces the need for large annotated
datasets. This method demonstrated improved Symbol Error Rates (SER)
when synthetic data complemented real samples, especially for the Borg
cipher manuscript dataset, making it valuable for OCR in historical
document processing,

For security applications, the paper \cite{parvez2020} addressed the generation
of handwritten Arabic CAPTCHAs using synthesized words. Their approach,
which incorporates segmentation-validation, provides enhanced protection
against automated recognition attacks compared to traditional CAPTCHAs.
With an efficiency of approximately 7 seconds per CAPTCHA and an
effectiveness rate of up to $63.43\%$, this method offers promising
applications for web security, digital archiving, and accessibility
tools.

\subsubsection{Deep Learning and Neural
Approaches}\label{deep-learning-and-neural-approaches}

Recent progress in deep learning has revolutionized handwriting
generation, with GANs playing a
central role. Study \cite{zheng2021} conducted a comparative study of various
GAN architectures---including Multi-Layer Perceptron (MLP) and
Convolutional Neural Network (CNN)--based GANs like DCGAN. The study
found that DCGAN, with its convolutional layers, Leaky ReLU activation,
and dropout regularization, offered superior image quality and training
efficiency, marking a significant advancement in GAN-based handwritten
data generation.

Enhancements in GAN methodologies continue to emerge. The method in
\cite{elanwar2024} examined Conditional GANs (CGANs) combined with recognition
networks and style banks, demonstrating improvements in Word Error Rate
(WER) and Character Error Rate (CER) on the IAM dataset---particularly
relevant for historical document processing. In another study, \cite{kang2020}
proposed GANwriting, which conditions text image generation on both
content and style using Adaptive Instance Normalization (AdaIN). This
technique produces high-quality, stylistically accurate text images,
though it struggles with capturing highly distinctive styles.

Other innovative approaches include the work of \cite{yuan2022}, which enhanced
GANs with skeleton information through a Self-attentive Refined
Attention Module (SAttRAM) for generating intricate brush handwriting
fonts. This method preserves structural integrity and manages geometric
variations, outperforming baseline models in content accuracy and FID
metrics \cite{yuan2022}. Study \cite{fogel2020} introduced ScrabbleGAN, a
semi-supervised model that synthesizes handwritten text images with
diverse styles and lexicons, significantly improving OCR performance
metrics such as WER and Normalized Edit Distance (NED) on the RIMES
dataset.

Hybrid models that merge adversarial generative techniques with
autoencoders have also shown promise. Research from \cite{wang2021} developed a
feature adversarial generative model integrated with an autoencoder to
generate handwritten Xibo characters. Despite the model's computational
complexity, its ability to produce visually coherent and stylistically
accurate fonts---validated by metrics like MAE and MSE---represents a
noteworthy innovation.

Conditional GANs have further been applied to specific tasks. Research
from \cite{elaraby2022} utilizes CGANs for few-shot handwritten character
recognition (HCR), generating synthetic samples to fine-tune pre-trained
models. This approach reduced overfitting and achieved a $99.36\%$
accuracy with VGG-16 on Latin datasets, though generating targeted
synthetic data remains time-intensive. Alternative strategies include
diffusion models; study \cite{aksan2018} introduced diffusion probabilistic
models for offline handwriting generation. These models simplify
training and bypass adversarial losses, albeit with limited diversity
compared to GANs.

Recurrent models also play a role in handwriting synthesis. As described
in \cite{madaan2022}, which is a comprehensive review covering RNNs, LSTMs,
Reinforcement Learning (RL), and GANs, noting that models like Graves'
LSTM and RL-based approaches (e.g., GAIL) effectively capture complex
sequence dynamics. Furthermore, study \cite{patil2022} combined Mixture Density
Networks (MDNs) with GANs and few-shot learning techniques such as
Model-Agnostic Meta-Learning (MAML) to achieve high-quality generation,
evidenced by improved FID and subjective ratings on the IAM-OnDB
dataset.

Transformer-based models have also been applied. Work in \cite{riaz2022}
explored a Conv-Transformer architecture for Urdu handwriting
recognition. By combining convolutional feature extraction with
transformer-based sequence modeling, the study achieved a Character
Error Rate (CER) of $5.31\%$, demonstrating significant improvements in
OCR performance despite requiring substantial training resources. Hybrid
frameworks, such as the Read-Write-Learn (RWL) framework proposed by
\cite{boteanu2023}, integrate self-learning with pre-trained models to enhance text
detection, recognition, and generation. Although computationally
intensive, RWL significantly reduces CER by incorporating language
models for pseudo-labeling and handwriting synthesis. Similarly,
approach in \cite{kotani2020} utilizes the Decoupled Style Descriptor (DSD)
model, which uses RNNs to independently encode character-level and
writer-level styles, achieving $89.38\%$ accuracy in writer classification
from a single word, despite high computational costs.

\subsubsection{Data Augmentation
Techniques}\label{data-augmentation-techniques}

Data augmentation remains a critical strategy for enhancing the quality
and diversity of handwriting datasets. study \cite{desousa2024} provides a
systematic review of augmentation methods for Handwritten Text
Recognition (HTR), covering Digital Image Processing, Transfer Learning,
and deep learning techniques such as GANs and Diffusion Models. Their
evaluation---using metrics like FID, Inception Score (IS), and Geometry
Score (GS)---highlights both the benefits and challenges (including high
computational costs and generalization issues) associated with these
methods.

Innovative augmentation strategies have also been introduced. Proposed
techniques introduced in \cite{shonenkov2021}, such as HandWritten Blots, which
simulate strikethrough text via Bezier curves, and StackMix, which
employs weakly supervised learning to generate synthetic handwritten
text. These methods demonstrated significant improvements, achieving a
CER of $1.73\%$ and a WER of $7.9\%$ on the BenthamR0 dataset. In parallel,
study \cite{sareen2024} combined traditional image augmentation techniques (e.g.,
noise injection, rotation, translation, scaling) with advanced GAN-based
methods to enhance machine learning models for Gurumukhi script
recognition. Their approach, validated through metrics such as accuracy,
precision, recall, and F1-score, reported accuracy rates up to $90.82\%$
on the GHWD dataset using CNN models---demonstrating its practical
utility in OCR systems, postal automation, and historical document
digitization.

The architectural foundations of these methodologies are illustrated in Figure 5, which compares four primary handwriting generation approaches: GAN-based, VAE-based, Transformer-based, and Diffusion-based architectures. Each architecture offers distinct advantages in terms of training stability, output quality, and computational efficiency.

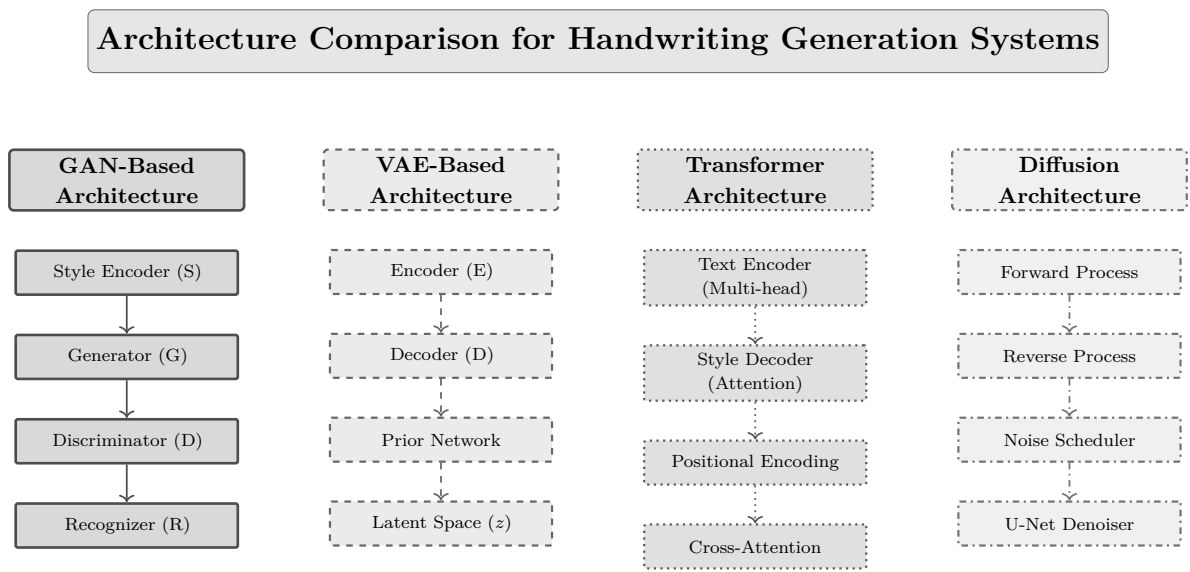
\begin{figure}[H]
\centering
\resizebox{0.95\textwidth}{!}{%
\begin{tikzpicture}[
    node distance=0.6cm,
    every node/.style={font=\small},
    title/.style={rectangle, rounded corners=3pt, fill=black!10, draw=black!60, 
                  text width=16cm, align=center, font=\Large\bfseries, minimum height=1cm},
    arch/.style={rectangle, rounded corners=2pt, fill=black!5, draw=black!60,
                 text width=3.5cm, align=center, font=\normalsize\bfseries, minimum height=0.8cm},
    component/.style={rectangle, rounded corners=1pt, fill=white, draw=black!50,
                      text width=3.3cm, align=center, font=\footnotesize, minimum height=0.7cm},
    gan/.style={fill=black!15, draw=black!70, line width=1.2pt},
    vae/.style={fill=black!8, draw=black!60, line width=1pt, dashed},
    transformer/.style={fill=black!12, draw=black!65, line width=1pt, dotted},
    diffusion/.style={fill=black!6, draw=black!55, line width=1pt, dashdotted}
]

\node[title] (title) {Architecture Comparison for Handwriting Generation Systems};

\node[arch, gan, below=1.2cm of title, xshift=-7.5cm] (gan-arch) {\textbf{GAN-Based}\\Architecture};
\node[component, gan, below=of gan-arch] (gan-style) {Style Encoder (S)};
\node[component, gan, below=of gan-style] (gan-gen) {Generator (G)};
\node[component, gan, below=of gan-gen] (gan-disc) {Discriminator (D)};
\node[component, gan, below=of gan-disc] (gan-rec) {Recognizer (R)};

\node[arch, vae, below=1.2cm of title, xshift=-2.5cm] (vae-arch) {\textbf{VAE-Based}\\Architecture};
\node[component, vae, below=of vae-arch] (vae-enc) {Encoder (E)};
\node[component, vae, below=of vae-enc] (vae-dec) {Decoder (D)};
\node[component, vae, below=of vae-dec] (vae-prior) {Prior Network};
\node[component, vae, below=of vae-prior] (vae-latent) {Latent Space ($z$)};

\node[arch, transformer, below=1.2cm of title, xshift=2.5cm] (trans-arch) {\textbf{Transformer}\\Architecture};
\node[component, transformer, below=of trans-arch] (trans-enc) {Text Encoder\\(Multi-head)};
\node[component, transformer, below=of trans-enc] (trans-dec) {Style Decoder\\(Attention)};
\node[component, transformer, below=of trans-dec] (trans-pos) {Positional Encoding};
\node[component, transformer, below=of trans-pos] (trans-cross) {Cross-Attention};

\node[arch, diffusion, below=1.2cm of title, xshift=7.5cm] (diff-arch) {\textbf{Diffusion}\\Architecture};
\node[component, diffusion, below=of diff-arch] (diff-forward) {Forward Process};
\node[component, diffusion, below=of diff-forward] (diff-reverse) {Reverse Process};
\node[component, diffusion, below=of diff-reverse] (diff-noise) {Noise Scheduler};
\node[component, diffusion, below=of diff-noise] (diff-unet) {U-Net Denoiser};

\draw[->, thick, black!70] (gan-style.south) -- (gan-gen.north);
\draw[->, thick, black!70] (gan-gen.south) -- (gan-disc.north);
\draw[->, thick, black!70] (gan-disc.south) -- (gan-rec.north);

\draw[->, thick, black!60, dashed] (vae-enc.south) -- (vae-dec.north);
\draw[->, thick, black!60, dashed] (vae-dec.south) -- (vae-prior.north);
\draw[->, thick, black!60, dashed] (vae-prior.south) -- (vae-latent.north);

\draw[->, thick, black!65, dotted] (trans-enc.south) -- (trans-dec.north);
\draw[->, thick, black!65, dotted] (trans-dec.south) -- (trans-pos.north);
\draw[->, thick, black!65, dotted] (trans-pos.south) -- (trans-cross.north);

\draw[->, thick, black!55, dashdotted] (diff-forward.south) -- (diff-reverse.north);
\draw[->, thick, black!55, dashdotted] (diff-reverse.south) -- (diff-noise.north);
\draw[->, thick, black!55, dashdotted] (diff-noise.south) -- (diff-unet.north);

\end{tikzpicture}%
}
\caption{Comparison of four handwriting generation architectures: GAN-based, VAE-based, Transformer-based, and Diffusion-based approaches.}
\label{fig:architecture_comparison}
\end{figure}

\subsection{The existing datasets for data augmentation and
generation}\label{the-existing-datasets-for-data-augmentation-and-generation}

This section gives a thorough overview of the most used datasets in
offline handwritten data augmentation and generation studies,
emphasizing their features, accessibility, and importance to the
research field.

One of the most extensively used datasets is the IAM Handwriting
Database, which has become a cornerstone in handwriting recognition
research. This dataset includes a vast collection of handwriting
samples, encompassing 1,539 pages produced by 657 different writers. It
features 63,000 words and 13,000 lines of text, offering a rich resource
for training and evaluating handwriting recognition systems. The dataset
is formatted as three-dimensional time series data and images, making it
versatile for various applications. It was created from teletype text
stories from the LOB Corpus and has been segmented into lines, words,
and characters, providing a detailed breakdown of handwritten English
text. Preprocessing steps such as binarization, size normalization,
skeletonization, and resizing to standard dimensions ensure that the
data is ready for use in recognition tasks. The IAM Handwriting Database
is publicly accessible, further contributing to its widespread adoption
in the research community. Study \cite{madaan2022} has not only utilized this
dataset in its research but have also provided detailed insights into
its structure and applications.

Similarly, the RIMES Dataset is another critical resource, particularly
in the context of French handwriting recognition. According to \cite{desousa2024},
this dataset comprises 12,723 text lines and 60,000 images collected
from 1,300 writers. The documents include forms and letters, making the
dataset highly relevant for automatic processing tasks. The images are
normalized and resized to a height of 32 pixels to maintain consistency
across the dataset. Like the IAM Handwriting Database, RIMES is publicly
available, which has led to its extensive use in the research community.

The CVL-Database provides another significant contribution, particularly
for those working with English handwritten documents. This dataset
includes 3,358 pages and 83,000 images from 311 participants, making it
one of the larger datasets in this domain. The CVL-Database is primarily
used for training and testing handwriting recognition systems, with the
images resized to a fixed height of 32 pixels for uniformity as it has
been utilized in study \cite{zdenek2021}.

The Bentham Dataset, as highlighted by \cite{shonenkov2021}, offers a unique
collection of handwritten pages by the British philosopher Jeremy
Bentham. With 25,000 pages of text, this dataset provides an invaluable
resource for historical document analysis and handwriting recognition.
The data is available as scanned images, capturing both historical
manuscripts and modern transcriptions, making it an essential resource
for researchers in this field.

the MNIST Database, as described in \cite{shao2020}, remains a foundational
resource. It consists of 60,000 training images and 10,000 test images
of handwritten digits. The images have been normalized to fit a $20 \times 20$
pixel box and are centered within a $28 \times 28$ image, providing a
standardized format that has been widely adopted in image processing
research.

Beyond English, multilingual handwriting datasets such as Omniglot and
MADCAT offer diverse resources for handwriting recognition across
different languages. Omniglot, described in \cite{elaraby2022}, includes
handwritten characters from 1,623 characters across various alphabets,
including Latin, Malay, Korean, and Sanskrit. This dataset is especially
useful for few-shot learning tasks due to its extensive variety of
characters. MADCAT, as highlighted in \cite{desousa2024}, on the other hand,
contains 750,000 images of mixed content lines, supporting document
analysis and recognition in multilingual contexts

The HKR Dataset focuses on handwritten text in Russian and Kazakh,
providing 64,943 lines of text in scanned image format. This dataset, as
highlighted by \cite{shonenkov2021}, is partially public, making it a valuable
resource for researchers interested in these languages. Similarly, the
Digital Peter Dataset offers 9,694 lines of handwritten Russian text,
available as scanned images and text files, further enriching the
resources available for multilingual handwriting recognition research.

In the context of Chinese handwriting, the CASIA Handwriting Database
according to \cite{shao2020} is a significant resource, containing 1.4 million
characters with 240 training samples and 60 test samples per class. The
characters are fitted to a $32 \times 32$ pixel box and centered in a $40 \times 40$
image, ensuring consistency and accuracy in character recognition tasks.
This dataset is publicly available, making it an essential tool for
researchers working on Chinese handwriting recognition. Another relevant
dataset is the Internal Calligraphy Set, as described in \cite{li2014}, it
focuses on strokes and characters derived from copybooks of the LIU
Gongquan style. Although this dataset is privately held, it provides
valuable insights into Chinese calligraphy and handwriting synthesis.

For Arabic handwriting, the IFN/ENIT Database is a key resource,
offering 26,459 binary images of handwritten Arabic words representing
Tunisian town and village names. Collected from 411 writers, this
dataset is publicly available for non-commercial research and is widely
used in Arabic handwriting recognition tasks \cite{abdi2015}. Additionally, the
AHCD (Arabic Handwritten Character Database) provides 16,800 images of
handwritten Arabic characters, making it another vital resource for
researchers \cite{mustapha2022}.

Urdu handwriting is also well-represented with datasets like the
NUST-UHWR Dataset, which contains 10,606 samples of handwritten text
lines, specifically designed for Urdu handwriting recognition.
Preprocessing steps include grayscale conversion and augmentation,
making it a comprehensive resource for this language. The UPTI-2 Dataset
and Ticker Dataset also contribute valuable data for Urdu text
recognition, with the former including one million samples and the
latter offering 19,437 samples of printed Urdu text as these Urdu
datasets have been utilized in study \cite{riaz2022}.

Other notable datasets include the OpenHaRT, DeepWriting Dataset, Brush
Handwriting Font Dataset, and Google Ngram Dataset, each contributing
unique data for handwriting recognition across various languages and
contexts. The OpenHaRT dataset, As documented in \cite{alonso2019}, has 710,892
images of handwritten words, shares similar preprocessing steps with the
RIMES dataset and is publicly available. The Brush Handwriting Font
Dataset highlighted by \cite{luo2022}, offers 15,799 high-resolution images
representing various calligraphy styles, and the Google Ngram Dataset
includes 176 n-grams derived from Google's Trillion Word
Corpus, used for handwritten text synthesis \cite{dey2016}.

These datasets collectively represent a wide array of resources
available for handwriting recognition, each contributing to the
advancement of offline handwritten data augmentation and generation
techniques. The detailed attributes of these datasets, along with their
preprocessing methods and public accessibility, make them invaluable
tools for researchers in this field.

Finally, Table 1 presents the distribution of all selected studies based
on the datasets and their respective languages.

\begin{longtable}{p{0.45\textwidth}p{0.38\textwidth}p{0.12\textwidth}}
\caption{Overview of the Selected Studies by Dataset, Including Type, Size, and Corresponding Languages}\\
\toprule
\textbf{Datasets and their languages} & \textbf{Size} & \textbf{References} \\
\midrule
\endfirsthead
\toprule
\textbf{Datasets and their languages} & \textbf{Size} & \textbf{References} \\
\midrule
\endhead
\bottomrule
\endlastfoot
IAM Handwriting Database (English) & 1,539 pages, 657 writers, 115,000
words & \cite{madaan2022} \\
RIMES Database (English) & 12,723 lines, 60,000 images & \cite{desousa2024} \\
CVL-Database (English) & 3,358 pages, 83,000 images & \cite{zdenek2021} \\
Bentham Dataset (English) & 25,000 pages & \cite{shonenkov2021} \\
MNIST Database (English) & 60,000 training, 10,000 test images &
\cite{shao2020} \\
Omniglot (Multilingual) & 1,623 characters, 20 samples each &
\cite{elaraby2022} \\
MADCAT (Multilingual) & 750,000 images & \cite{desousa2024} \\
CASIA Handwriting Database (Chinese) & 1.4 million characters &
\cite{shao2020} \\
Internal Calligraphy Set (Chinese) & 106 strokes & \cite{li2014} \\
IFN/ENIT Database (Arabic) & 26,459 images & \cite{abdi2015} \\
AHCD Database (Arabic) & 16,800 images & \cite{mustapha2022} \\
NUST-UHWR Dataset (Urdu ) & 10,606 text lines & \cite{riaz2022} \\
UPTI-2 Dataset (Urdu) & One million samples & \cite{riaz2022} \\
OpenHaRT (Arabic) & 710,892 images & \cite{alonso2019} \\
Brush Handwriting Font Dataset (Multilingual) & 15,799 images &
\cite{luo2022} \\
Google Ngram Dataset (English) & 176 n-grams & \cite{dey2016} \\
\end{longtable}

\subsection{Evaluation Metrics and
Benchmarks}\label{evaluation-metrics-and-benchmarks}

The evaluation of offline handwritten data augmentation and generation
techniques is essential for ensuring quality and usability in tasks like
optical character recognition (OCR). Metrics are categorized into
\textbf{Quantitative Metrics} and \textbf{Qualitative Assessments}, each
assessing different aspects of the generated data.

\textbf{Quantitative Metrics}

These metrics provide numerical evaluations of model performance. Error metrics such as Mean Absolute Error (MAE) and Mean Squared Error (MSE) measure prediction accuracy \cite{wang2021}. Text recognition metrics including CER and WER assess transcription accuracy \cite{boteanu2023}, \cite{shonenkov2021}. Performance metrics such as Accuracy and F1 Score measure prediction correctness and balance between precision and recall \cite{elarian2015}. Generalization Error (Etest) evaluates performance on unseen data \cite{elaraby2022}. Image quality metrics including Frechet Inception Distance (FID) and Geometry Score (GS) assess visual realism of generated handwriting \cite{pippi2023}. Edit distance metrics such as Levenshtein Distance measure the similarity between generated and real text \cite{alonso2019}. Additionally, specialized metrics including Probability for Target Shifts and Binary Cross-Entropy Losses evaluate spatial and temporal handwriting features \cite{kotani2020}.

\begin{longtable}{p{0.38\textwidth}p{0.58\textwidth}}
\caption{Summary Table of Quantitative Metrics Used Across Studies}\\
\toprule
\textbf{Quantitative Metric} & \textbf{Studies Utilizing the Metric} \\
\midrule
\endfirsthead
\toprule
\textbf{Quantitative Metric} & \textbf{Studies Utilizing the Metric} \\
\midrule
\endhead
\bottomrule
\endlastfoot
Mean Absolute Error (MAE) & \cite{wang2021} \\
Mean Squared Error (MSE) & \cite{wang2021} \\
Character Error Rate (CER) & \cite{boteanu2023}, \cite{shonenkov2021} \\
Accuracy & \cite{boteanu2023}, \cite{zheng2021}, \cite{elarian2015} \\
F1 Score & \cite{abdi2015}, \cite{elaraby2022}, \cite{elarian2015} \\
Generalization Error & \cite{elaraby2022} \\
Frechet Inception Distance (FID) & \cite{alonso2019}, \cite{pippi2023}, \cite{kang2020icfhr} \\
Geometry Score (GS) & \cite{pippi2023} \\
Levenshtein Distance (Edit Distance) & \cite{alonso2019}, \cite{shonenkov2021strikethrough} \\
Word Error Rate (WER) & \cite{alonso2019}, \cite{shonenkov2021strikethrough} \\
Probability for Target Shifts ($\Delta x$, $\Delta y$) & \cite{kotani2020} \\
Binary Cross-Entropy Losses (eos, eoc) & \cite{kotani2020} \\
\end{longtable}

\textbf{Qualitative Assessments}

These involve subjective evaluations of the visual quality and realism of generated handwriting. Visual inspection involves experts assessing the authenticity of generated fonts \cite{wang2021}. Human-based assessments employ annotators to evaluate the realism and diversity of handwriting samples \cite{kang2020icfhr}. User preference studies require participants to compare handwriting samples and select the most realistic \cite{kotani2020}.

\textbf{Evaluation Protocols}

Researchers use \textbf{Comparative Analysis Protocols}, combining both
quantitative and qualitative metrics, often benchmarking against
standard datasets like the IAM dataset \cite{pippi2023}.

This comprehensive evaluation framework, integrating numerical accuracy
and visual authenticity, provides valuable insights into the strengths
and limitations of handwritten data generation techniques, guiding
future advancements in the field.

\section{Challenges and Techniques in Handwritten
Generation}\label{challenges-and-techniques-in-handwritten-generation}

\subsection{Challenges}\label{challenges}

Generating offline handwritten text comes with a lot of challenges that
researchers and developers need to tackle to make models more accurate,
diverse, and useful. From capturing different handwriting styles to
dealing with technical limitations, there are many hurdles to overcome.
Here are some of the biggest challenges in handwriting generation.

One of the toughest challenges is style variability and realism.
Everyone's handwriting is different, and current models
often struggle to capture all these variations, especially when faced
with styles they haven't seen before. This can lead to
overfitting, where the model gets too focused on a small set of styles
and becomes less flexible. Another issue is finding the right balance
between quality and diversity—sometimes models generate neat
handwriting but lack variety in styles \cite{luo2022}.

Another big problem is dataset bias and generalization. Most handwriting
generation models are trained on datasets that only cover a limited
number of languages and styles. This makes it hard for them to adapt to
new scripts, especially for languages that don't have a lot of
handwritten data available \cite{yuan2022}, \cite{mattick2021}. This issue is especially
serious in low-resource settings, where models need to work with minimal
training data \cite{kang2022}.

Speaking of low-resource languages, data scarcity is a major roadblock.
Unlike widely used languages like English, many other languages don't
have large, well-labeled handwriting datasets. Collecting handwritten
samples can be difficult, especially in places with lower literacy rates
or fewer contributors \cite{madaan2022}. Without enough data, it's
tough to train effective models, which means researchers have to come up
with new ways to work around this limitation \cite{davis2020}.

Then there's linguistic diversity. Many languages come with their own
unique challenges, such as different scripts, complicated spelling
rules, and the heavy use of diacritics. This makes handwriting
generation even harder since every script or dialect might need its own
specialized model \cite{kotani2020}. On top of that, handwriting styles naturally
vary across languages, making the task even trickier \cite{mustapha2022}.

Computational constraints also pose a big challenge. Training
handwriting generation models---especially advanced ones like GANs and
transformers---requires a lot of computing power. This can limit both
research and real-world applications, especially in places where
high-end hardware isn't available \cite{zheng2021}. Even when
resources are available, generating and processing handwriting data
takes up a lot of computing power, making it less accessible \cite{mattick2021}.

There are also technical limitations to consider. Many handwriting
models rely on predefined alphabets, which don't always work well for
complex scripts. For example, Sequential-to-Sequential (Seq2Seq) models
often struggle with languages that have intricate writing systems,
meaning they need significant adjustments to perform well \cite{liu2021}.

Another issue with GAN-based models is training stability and mode
collapse. GANs can be unstable during training, and sometimes they get
stuck generating the same few handwriting samples over and over again,
rather than creating a variety of different styles \cite{kang2020}.

On top of that, handling extreme variations in handwriting is still a
big challenge. Some people write in very cursive or artistic ways, which
can be difficult for models to reproduce accurately \cite{lian2016}.

Lastly, there's the issue of bias in handwriting generation. Many models
are trained on datasets that aren't fully representative of all writing
styles and languages, which can lead to biased outputs. Some dialects or
styles might be overrepresented while others are ignored. This can make
models less fair and less useful for a wider range of users \cite{jiang2017}.

Tackling these challenges is key to making handwriting generation better
and more widely usable. Researchers need to focus on building more
diverse datasets, improving model architectures, and finding ways to
make these models more efficient so that they can work well even in
low-resource environments.

\subsection{Techniques}\label{techniques}

One of the biggest hurdles in handwriting generation is the lack of
sufficient training data, especially for low-resource languages. To
tackle this, researchers use different techniques to artificially
increase the size and diversity of available datasets. These methods
ensure that models can learn from a wider range of handwriting styles,
making them more adaptable and effective.

A common approach is data augmentation, where techniques like rotation,
scaling, contrast adjustment, and GAN-based augmentation are used to
expand existing datasets. These techniques help generate a variety of
handwriting samples, making the training process more robust \cite{sareen2024}.
One specific method, called Mixup, combines two different handwriting
samples to create synthetic data points, improving dataset size and
diversity \cite{wang2021}. Another technique, Elastic Distortion, introduces
non-uniform distortions in handwriting samples to better simulate
natural variations. Similarly, Random Perturbations, which involve small
changes like shifting or rotating text, make datasets richer by adding
new variations. To ensure that essential handwriting features are not
lost during augmentation, Attention-Based Character Localization helps
maintain the quality and clarity of the generated samples \cite{mustapha2022}.

Another approach is Transfer learning, which has proven to be a highly
effective method for handling data scarcity. Instead of training a model
from scratch, researchers use pre-trained models that have already
learned from large datasets in high-resource languages and then
fine-tune them using smaller datasets from low-resource languages. This
process helps models adapt to new scripts with significantly less
training data \cite{pippi2023}. For example, Text and Style Conditioned GANs
trained on English handwriting can be fine-tuned for other languages,
allowing them to generate handwriting with specific stylistic and
content-based constraints \cite{davis2020}.

In addition to data-focused solutions, developing specialized model
architectures plays a crucial role in improving handwriting generation.
Some advanced architectures, such as Multilingual-GANs, are designed to
convert printed text images into handwritten text without relying on
predefined alphabets. This makes them highly adaptable for different
languages without requiring extensive retraining \cite{huu2021}. Many of these
models also incorporate multi-task learning and auxiliary classifiers,
which help them better capture the unique characteristics of various
scripts. For instance, generating Arabic handwriting requires special
attention to diacritical marks, which can completely change the meaning
of words. By incorporating these features into the architecture, models
can ensure that the generated handwriting is both accurate and visually
natural (Mustapha et al., 2021). Some researchers have also combined
CNNs, RNNs, and GANs to create hybrid models that improve both content
accuracy and stylistic realism, making them better suited for handling
script variations \cite{mattick2021}.

Another promising technique is cross-lingual transfer, where models
trained on multiple languages are used to improve handwriting generation
in low-resource languages. By learning from multilingual datasets, these
models can identify and apply patterns across different scripts, making
them more effective when dealing with languages that have limited
handwriting data \cite{liu2021}. This approach works particularly well when
the target language has similarities with a high-resource language,
allowing the model to transfer relevant features more effectively
\cite{mattick2021}. One technique called StackMix helps enhance dataset quality by
identifying character boundaries and generating synthetic handwritten
text from isolated characters, further improving handwriting generation
for low-resource languages \cite{shonenkov2021strikethrough}.

By leveraging these techniques---data augmentation, transfer learning,
specialized model architectures, and cross-lingual
transfer---researchers are making significant progress in overcoming
data limitations. These approaches help make handwriting generation
models more efficient, adaptable, and capable of handling a diverse
range of languages and scripts.

\section{Discussion}\label{discussion}

This section presents a detailed analysis of the findings, where each of
the research questions are addressed, and insights into the evolution of
methods, their effectiveness, challenges, and the potential for future
advancements are provided in this field.

\textbf{RQ1:} How have offline handwritten data augmentation and
generation techniques evolved, and what are the current state-of-the-art
methods?

The evolution of offline handwritten data augmentation and generation
techniques has been marked by the transition from simple geometric
transformations to more sophisticated deep learning models. Initially,
techniques like random elastic deformations and geometric
transformations were prevalent but often insufficient in capturing the
variability of human handwriting \cite{madaan2022}. The advent of Generative
Adversarial Networks (GANs) revolutionized the field, enabling the
generation of high-quality, realistic handwritten text images \cite{zheng2021}.
CGANs and diffusion models are more recent
advancements, offering improved performance and flexibility in
generating diverse handwriting styles \cite{aksan2018}. These methods represent
the current state-of-the-art, with GANs being the most widely adopted
due to their ability to synthesize realistic and stylistically diverse
handwriting \cite{fogel2020}.

\textbf{RQ2:} How do these techniques compare in terms of effectiveness
and efficiency for enhancing OCR accuracy and overall recognition
performance?

In comparing the effectiveness and efficiency of these techniques, GANs,
particularly CGANs, have demonstrated superior
performance in enhancing OCR accuracy and overall recognition
performance. Studies have shown that integrating GAN-generated data into
training datasets significantly reduces WER and
CER, especially when applied to low-resource
languages \cite{elanwar2024}. However, these models often require substantial
computational resources, which can be a limiting factor in their
widespread adoption. Diffusion models, while less computationally
intensive, offer a trade-off between quality and efficiency, making them
a viable alternative in scenarios where computational resources are
constrained \cite{aksan2018}. Traditional data augmentation methods like
geometric transformations remain relevant for their simplicity and low
computational cost, though they fall short in terms of the diversity and
realism of generated samples compared to deep learning approaches
\cite{shonenkov2021strikethrough}. Figure 6 provides visual examples of synthetic handwritten text generated by different models, demonstrating the substantial improvements in quality and diversity achieved by modern synthesis techniques.

\begin{figure}[H]
\centering
\begin{subfigure}[b]{0.48\textwidth}
    \centering
    \includegraphics[width=\textwidth]{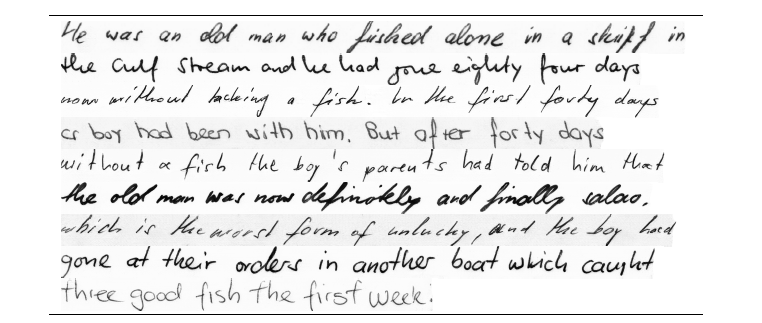}
    \caption{Content and style aware synthesis output}
    \label{fig:content_style_aware}
\end{subfigure}
\hfill
\begin{subfigure}[b]{0.48\textwidth}
    \centering
    \includegraphics[width=\textwidth]{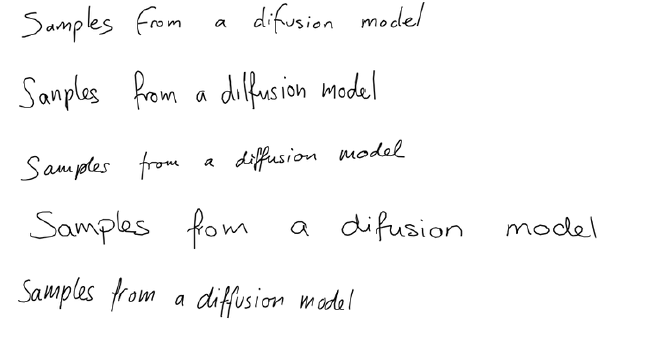}
    \caption{Diffusion model generated handwriting}
    \label{fig:diffusion_handwriting}
\end{subfigure}

\vspace{0.5cm}

\begin{subfigure}[b]{0.31\textwidth}
    \centering
    \includegraphics[width=\textwidth]{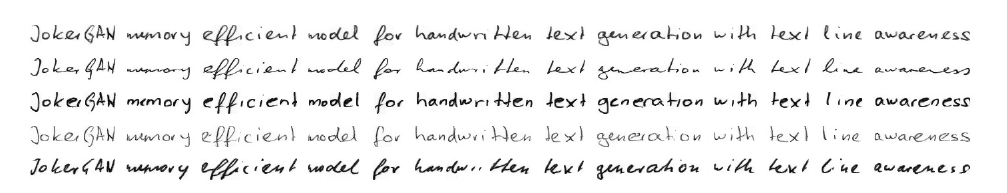}
    \caption{JokerGAN generated samples}
    \label{fig:jokergan}
\end{subfigure}
\hfill
\begin{subfigure}[b]{0.31\textwidth}
    \centering
    \includegraphics[width=\textwidth]{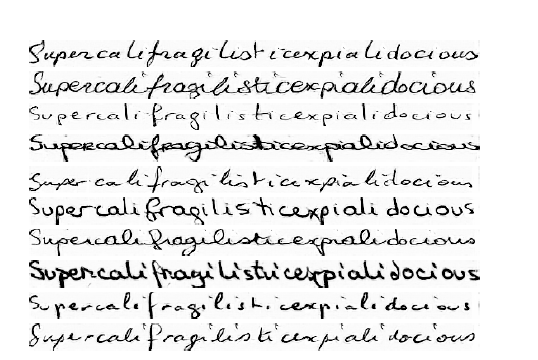}
    \caption{ScrabbleGAN generated text}
    \label{fig:scrabblegan}
\end{subfigure}
\hfill
\begin{subfigure}[b]{0.31\textwidth}
    \centering
    \includegraphics[width=\textwidth]{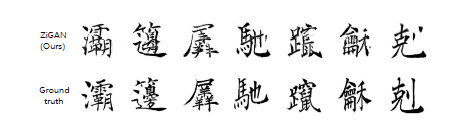}
    \caption{ZIGAN generated handwriting}
    \label{fig:zigan}
\end{subfigure}

\caption{Examples of synthetic handwritten text generated by different models: (a,b) show outputs from content-style aware and diffusion-based approaches, while (c,d,e) present handwriting samples generated by JokerGAN, ScrabbleGAN, and ZIGAN respectively, demonstrating the quality and diversity achievable by modern handwriting synthesis techniques.}
\label{fig:handwriting_synthesis_techniques}
\end{figure}

Our systematic analysis reveals clear trends in methodological preferences within the field. Figure 7 shows the distribution of data augmentation approaches used across the surveyed studies, highlighting the dominance of GAN-based methods, which account for over 60\% of the research efforts.

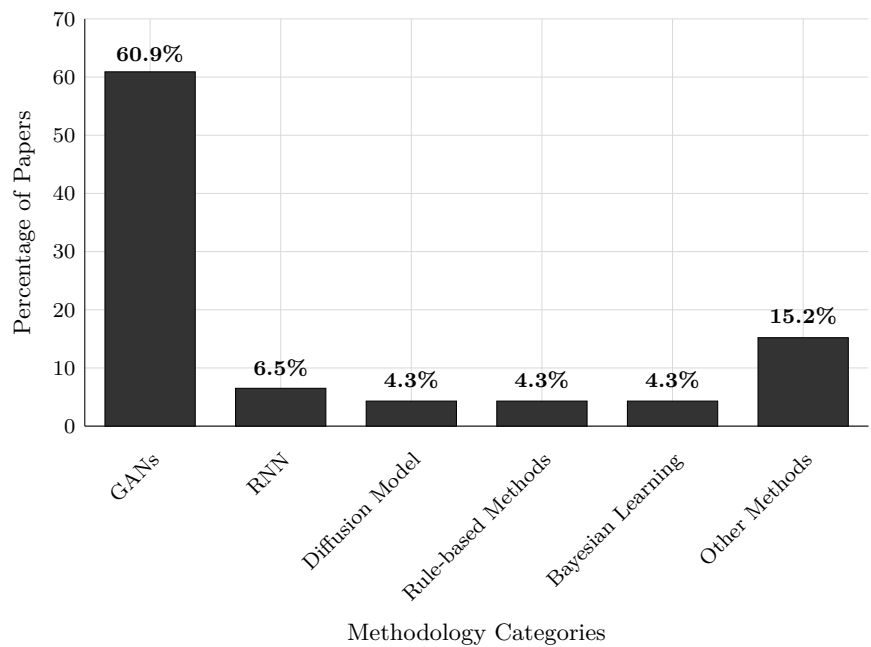
\begin{figure}
\centering
\begin{tikzpicture}
\begin{axis}[
    ybar,
    width=12cm,
    height=7cm,
    ymin=0,
    ymax=70,
    ylabel={Percentage of Papers},
    xlabel={Methodology Categories},
    ylabel style={font=\small},
    xlabel style={font=\small},
    xtick=data,
    xticklabels={GANs, RNN, Diffusion Model, Rule-based Methods, Bayesian Learning, Other Methods},
    xticklabel style={rotate=45, anchor=north east, font=\footnotesize},
    ytick={0,10,20,30,40,50,60,70},
    yticklabel style={font=\footnotesize},
    bar width=1.2cm,
    enlarge x limits=0.1,
    grid=major,
    grid style={line width=0.1pt, draw=gray!30},
    axis lines*=left,
    tick style={draw=none},
    every axis plot/.append style={fill},
]

\addplot[fill=black!80, draw=black] coordinates {
    (1,60.9)
    (2,6.5)
    (3,4.3)
    (4,4.3)
    (5,4.3)
    (6,15.2)
};

\node at (axis cs:1,64) {\footnotesize\textbf{60.9\%}};
\node at (axis cs:2,10) {\footnotesize\textbf{6.5\%}};
\node at (axis cs:3,8) {\footnotesize\textbf{4.3\%}};
\node at (axis cs:4,8) {\footnotesize\textbf{4.3\%}};
\node at (axis cs:5,8) {\footnotesize\textbf{4.3\%}};
\node at (axis cs:6,19) {\footnotesize\textbf{15.2\%}};

\end{axis}
\end{tikzpicture}
\caption{Proportion of data augmentation approaches used by studies. Each work may have more than one approach associated}
\end{figure}

Similarly, Figure 8 presents the distribution of datasets utilized in the research, with the IAM dataset being the most frequently used resource, appearing in nearly 20\% of the studies.

\begin{figure}
\centering
\begin{tikzpicture}
\begin{axis}[
    ybar,
    width=12cm,
    height=7cm,
    ymin=0,
    ymax=25,
    ylabel={Usage Percentage},
    xlabel={Handwriting Datasets},
    ylabel style={font=\small},
    xlabel style={font=\small},
    xtick=data,
    xticklabels={IAM, RIMES, Saint Gall, HKR, Digital Peter},
    xticklabel style={rotate=45, anchor=north east, font=\footnotesize},
    ytick={0,5,10,15,20,25},
    yticklabel style={font=\footnotesize},
    bar width=1.2cm,
    enlarge x limits=0.1,
    grid=major,
    grid style={line width=0.1pt, draw=gray!30},
    axis lines*=left,
    tick style={draw=none},
    every axis plot/.append style={fill},
]

\addplot[fill=black!80, draw=black] coordinates {
    (1,19.4)
    (2,8.3)
    (3,5.6)
    (4,5.6)
    (5,5.6)
};

\node at (axis cs:1,21) {\footnotesize\textbf{19.4\%}};
\node at (axis cs:2,10) {\footnotesize\textbf{8.3\%}};
\node at (axis cs:3,7.5) {\footnotesize\textbf{5.6\%}};
\node at (axis cs:4,7.5) {\footnotesize\textbf{5.6\%}};
\node at (axis cs:5,7.5) {\footnotesize\textbf{5.6\%}};

\end{axis}
\end{tikzpicture}
\caption{Proportion of datasets used by studies. Each work may have more than one dataset associated with}
\end{figure}
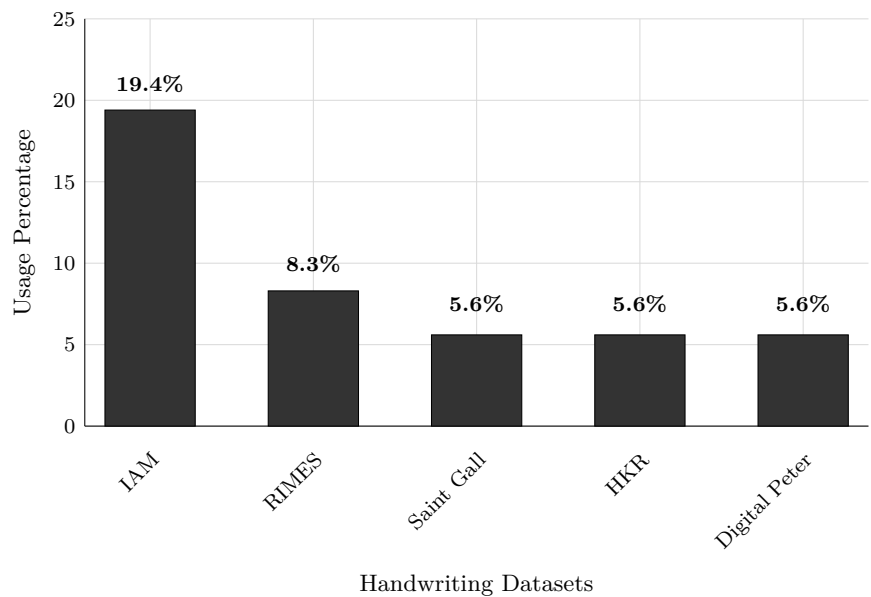

\textbf{RQ3:} What are the main challenges in creating realistic and
diverse synthetic handwriting samples, and how can they be addressed?

Creating realistic and diverse synthetic handwriting samples presents
several challenges, including the high variability of handwriting
styles, the need to preserve the stylistic features of specific scripts,
and the risk of mode collapse in GANs. The variability in handwriting,
particularly for complex scripts and low-resource languages, requires
models that can capture subtle stylistic nuances while maintaining
realism. One of the main challenges is the computational cost associated
with training GANs, as well as the instability and mode collapse issues
that can occur during training \cite{fogel2020}. To address these challenges,
researchers have explored advanced GAN architectures, such as StyleGAN,
which offers better style-content disentanglement, and diffusion models,
which provide a more stable training process \cite{davis2020}. Transfer learning
and data augmentation techniques, including Elastic Distortion and
Mixup, have also been employed to enhance diversity and improve model
performance in low-resource scenarios \cite{pippi2023}.

\textbf{RQ4:} How can these techniques be utilized to increase and
diversify datasets, particularly for low-resource languages?

For low-resource languages, the application of these techniques is
crucial in overcoming the challenges posed by limited annotated data.
Transfer learning allows models pre-trained on high-resource languages
to be fine-tuned on smaller datasets, thereby improving performance
without the need for extensive training data \cite{elaraby2022}. GANs and CGANs
can generate diverse handwriting samples that enrich training datasets,
increasing their size and variability \cite{fogel2020}. Additionally, data
augmentation techniques like rotation, scaling, and GAN-based approaches
are particularly effective in creating synthetic data that reflects the
diversity of handwriting styles within low-resource languages \cite{sareen2024}.
These techniques not only expand the dataset but also help in preserving
the linguistic and stylistic nuances of low-resource languages, making
them more accessible for OCR and other handwriting recognition systems.

\textbf{RQ5:} Which datasets are most used and most effective for
offline handwritten data augmentation and generation?

The IAM Handwriting Database and RIMES Dataset are among the most
commonly used and effective datasets for offline handwritten data
augmentation and generation, primarily due to their extensive
collections of annotated handwritten text which have been utilized by
researches as \cite{madaan2022}, and \cite{desousa2024}. These datasets provide a wide
variety of handwriting samples, which are essential for training and
evaluating handwriting recognition systems. The CVL-Database and Bentham
Dataset are also widely used, offering rich resources for English
handwriting recognition and historical manuscript analysis, respectively
\cite{zdenek2021}. For multilingual contexts, datasets like Omniglot and MADCAT
offer valuable resources for handwriting recognition across different
languages \cite{shao2020}, while CASIA and IFN/ENIT are critical for Chinese
and Arabic handwriting, respectively \cite{abdi2015}. These datasets have become
benchmarks in the field, driving advancements in handwriting recognition
technologies.

\textbf{RQ6:} How can the quality and diversity of augmented and
generated handwriting be evaluated, especially for languages with
limited existing datasets?

The quality and diversity of augmented and generated handwriting can be
evaluated using both quantitative metrics and qualitative assessments.
Quantitative metrics such as Frechet Inception Distance (FID), Character
CER, and WER are commonly used to assess
the similarity and accuracy of generated handwriting compared to real
samples \cite{alonso2019}. Additionally, Levenshtein Distance and Mean Squared
Error (MSE) provide insights into the structural consistency and
precision of the generated text \cite{aksan2018}. For languages with limited
existing datasets, qualitative assessments, including visual inspection
by experts and human-based evaluations, are crucial in ensuring that the
generated handwriting maintains the stylistic integrity and diversity
necessary for effective recognition. Combining these evaluation methods
provides a comprehensive framework for assessing the effectiveness of
augmentation and generation techniques, ensuring that they meet the
desired standards for both quality and diversity.

\section{Emerging Techniques and Future
Directions}\label{emerging-techniques-and-future-directions}

Handwriting generation is constantly evolving, with new techniques
emerging that promise to improve quality, adaptability, and performance.
As research progresses, these advancements are helping overcome existing
challenges while paving the way for more efficient and diverse
handwriting models.

\subsection{Emerging Techniques}\label{emerging-techniques}

Recent breakthroughs in generative modeling have introduced several
promising approaches:

Diffusion Models have shown impressive results in generating
high-quality and diverse handwritten text images. These models are
becoming popular due to their ability to quickly adapt to different
writing styles and perform well across various tasks \cite{patil2022}.

Vision Transformer (ViT) Models are gaining attention, particularly in
tasks involving few-shot handwriting character recognition (HCR).
Compared to traditional CNNs, ViTs offer better generalization, making
them more effective in recognizing diverse handwriting styles \cite{elaraby2022}.

Improved GAN Variations such as StyleGAN have been successful in
separating writing style from content, enabling greater flexibility in
handwriting generation.Also, more advanced Progressive GANs have also
been shown to enhance the realism and consistency of synthesized
handwriting \cite{davis2020}.

Integration of Data Augmentation with Data Synthesis has proven
effective in expanding handwriting datasets. By combining traditional
augmentation techniques with generative models, researchers are
producing higher-quality and more varied training data, which improves
handwriting recognition systems \cite{mattick2021}.

Self-Learning Frameworks are opening new possibilities for handwriting
generation. By using language models, these frameworks allow for better
style adaptation without requiring large labeled datasets. Similarly,
self-supervised and semi-supervised learning approaches leverage large
amounts of unlabeled data to enhance model robustness and versatility
\cite{ding2023}.

Zero-Shot and Continual Learning are also gaining traction. Zero-shot
learning allows handwriting models to generate text in styles they have
never seen before, while continual learning helps models adapt to
evolving data over time \cite{zheng2021}.

\subsection{Future Directions}\label{future-directions}

Future handwriting generation research should focus on several related areas that can improve the field's capabilities and reach. First, developing better neural network designs is essential. These improved models should use refined activation functions in GANs and apply large-scale pre-training followed by focused fine-tuning. Such methods can significantly improve model performance across different handwriting styles and languages \cite{riaz2022}.

Building on these improvements, adaptive data augmentation represents another important direction. Future systems should automatically adjust their augmentation methods based on the specific features of different handwriting styles and scripts. This flexibility would create stronger systems that can easily handle the specific needs of various writing traditions \cite{pippi2023}. Combining these adaptive techniques with pre-trained language models shows great promise, as these models can greatly improve the quality and consistency of generated text, especially for multiple languages and scripts \cite{ding2023}.

The field also needs broader collaboration beyond computer science. Working with cognitive scientists, linguists, and historians can provide valuable insights into how handwriting works, leading to better feature extraction and annotation methods. These partnerships can improve both generation capabilities and classification techniques, creating a more complete understanding of handwriting as both a thinking and cultural process \cite{haines2016}.

At the same time, practical implementation requires attention to computational efficiency. Future research must develop efficient model designs using techniques like pruning and quantization to balance high performance with accessibility. This is crucial for making handwriting generation models practical for real-world use, especially in environments with limited computing power \cite{kang2020}.

Most importantly, the future of handwriting generation depends on including more languages and cultural writing styles. Research must focus on creating datasets that cover a wider range of languages and handwriting traditions. This is essential for ensuring that handwriting generation systems work well for diverse communities worldwide \cite{davis2020}.

By exploring these connected research areas, handwriting generation technology can reach new levels of adaptability, efficiency, and accessibility. The combination of better architectures, adaptive methods, interdisciplinary insights, efficient computing, and cultural diversity will drive the field toward more advanced and widely useful solutions.

\section{Conclusion}\label{conclusion}

Offline handwritten data augmentation and generation techniques have
emerged as pivotal elements in advancing Handwritten Text Recognition
(HTR) systems. This survey provided an in-depth examination of the
various methods employed to enhance HTR systems through synthetic data
generation. Our main contributions include defining the scope of this
survey to focus on offline handwritten text generation, exploring the
datasets and methods used to synthesize handwriting images, and
analyzing the evolution of these techniques over the past decade. We
also identified current gaps and challenges in the literature, which
have guided our suggestions for future research directions.

In this study, we began by collecting a substantial number of relevant
academic papers, which were meticulously filtered through a systematic
exclusion process. Ultimately, a curated selection of studies was
reviewed in detail, allowing us to map the most commonly used datasets
and recognition levels in the field of handwritten text generation. This
mapping enabled us to contextualize each method within its specific
application domain, providing a clearer understanding of its
effectiveness and potential.

Our findings indicate that traditional digital image processing
techniques, while still valuable, are increasingly being supplemented or
even replaced by advanced generative models such as Generative
Adversarial Networks (GANs). GANs have shown considerable promise in
generating realistic handwritten text images, and their integration with
optical models is poised to be a significant trend in the coming years.
The use of GANs represents a shift towards more sophisticated methods
that can produce high-quality synthetic data, potentially transforming
how HTR systems are trained and optimized.

It is worth noting that the field of offline handwritten text
recognition, particularly with a focus on data augmentation, is still
relatively nascent. However, recent advancements, especially in the
application of GANs, suggest a growing interest and substantial progress
within the academic community. This trend underscores the increasing
recognition of the benefits that generative models can bring to HTR,
particularly in enhancing the robustness and accuracy of these systems.

In conclusion, future research should consider the application of these
techniques to low-resource scenarios, where the generation of synthetic
handwriting images can play a critical role in training optical models
effectively. Additionally, the development of generative models that are
closely integrated with optical recognition systems, possibly within a
self-supervised learning framework, represents a promising direction for
future studies. These advancements have the potential to significantly
improve the performance of HTR systems, making them more adaptable and
accurate across various languages and scripts.


\clearpage
\bibliography{references}

\begin{thebibliography}{10}

\bibitem{boteanu2023}
A.~Boteanu, D.~Cheng, and S.~Kadioğlu.
\newblock Read-write-learn: Self-learning for handwriting recognition.
\newblock In {\em Proceedings of the ACM Symposium on Document Engineering
  2023}, pages 1--4, Limerick, Ireland, Aug 2023. ACM.

\bibitem{madaan2022}
M.~Madaan, A.~Kumar, S.~Kumar, A.~Saha, and K.~Gupta.
\newblock Handwriting generation and synthesis: A review.
\newblock In {\em 2022 Second International Conference on Power, Control and
  Computing Technologies (ICPC2T)}, pages 1--6, Raipur, India, Mar 2022. IEEE.

\bibitem{aksan2018}
E.~Aksan, F.~Pece, and O.~Hilliges.
\newblock Deepwriting: Making digital ink editable via deep generative
  modeling.
\newblock In {\em Proceedings of the 2018 CHI Conference on Human Factors in
  Computing Systems}, pages 1--14, Montreal QC, Canada, Apr 2018. ACM.

\bibitem{moher2009}
D.~Moher.
\newblock Preferred reporting items for systematic reviews and meta-analyses:
  The prisma statement.
\newblock {\em Annals of Internal Medicine}, 2009.

\bibitem{kitchenham2009}
B.~Kitchenham, O.~Pearl~Brereton, D.~Budgen, M.~Turner, J.~Bailey, and
  S.~Linkman.
\newblock Systematic literature reviews in software engineering -- a systematic
  literature review.
\newblock {\em Inf. Softw. Technol.}, 51(1):7--15, Jan 2009.

\bibitem{kitchenham2010}
B.~Kitchenham et~al.
\newblock Systematic literature reviews in software engineering -- a tertiary
  study.
\newblock {\em Inf. Softw. Technol.}, 52(8):792--805, Aug 2010.

\bibitem{abdi2015}
M.~N. Abdi and M.~Khemakhem.
\newblock A model-based approach to offline text-independent arabic writer
  identification and verification.
\newblock {\em Pattern Recognit.}, 48(5):1890--1903, May 2015.

\bibitem{li2014}
W.~Li, Y.~Song, and C.~Zhou.
\newblock Computationally evaluating and synthesizing chinese calligraphy.
\newblock {\em Neurocomputing}, 135:299--305, Jul 2014.

\bibitem{jiang2017}
Y.~Jiang, Z.~Lian, Y.~Tang, and J.~Xiao.
\newblock Dcfont: an end-to-end deep chinese font generation system.
\newblock In {\em SIGGRAPH Asia 2017 Technical Briefs}, pages 1--4, Bangkok,
  Thailand, Nov 2017. ACM.

\bibitem{dey2016}
A.~U. Dey and G.~Harit.
\newblock Generating synthetic handwriting using n-gram letter glyphs.
\newblock In {\em Proceedings of the Tenth Indian Conference on Computer
  Vision, Graphics and Image Processing}, pages 1--8, Guwahati Assam, India,
  Dec 2016. ACM.

\bibitem{balreira2017}
D.~G. Balreira and M.~Walter.
\newblock Handwriting synthesis from public fonts.
\newblock In {\em 2017 30th SIBGRAPI Conference on Graphics, Patterns and
  Images (SIBGRAPI)}, pages 246--253, Niteroi, Oct 2017. IEEE.

\bibitem{souibgui2021}
M.~A. Souibgui et~al.
\newblock One-shot compositional data generation for low resource handwritten
  text recognition.
\newblock {\em arXiv}, Oct 2021.
\newblock arXiv:2105.05300.

\bibitem{parvez2020}
M.~T. Parvez and S.~A. Alsuhibany.
\newblock Segmentation-validation based handwritten arabic captcha generation.
\newblock {\em Comput. Secur.}, 95:101829, Aug 2020.

\bibitem{zheng2021}
S.~Zheng.
\newblock Analysis of generating handwriting based on gan model with different
  structures.
\newblock In {\em 2021 IEEE International Conference on Computer Science,
  Electronic Information Engineering and Intelligent Control Technology (CEI)},
  pages 654--658, Fuzhou, China, Sep 2021. IEEE.

\bibitem{elanwar2024}
R.~Elanwar and M.~Betke.
\newblock Generative adversarial networks for handwriting image generation: a
  review.
\newblock {\em Vis. Comput.}, Jul 2024.

\bibitem{kang2020}
L.~Kang, P.~Riba, Y.~Wang, M.~Rusiñol, A.~Fornés, and M.~Villegas.
\newblock Ganwriting: Content-conditioned generation of styled handwritten word
  images.
\newblock {\em arXiv}, Jul 2020.
\newblock arXiv:2003.02567.

\bibitem{yuan2022}
S.~Yuan, R.~Liu, M.~Chen, B.~Chen, Z.~Qiu, and X.~He.
\newblock Se-gan: Skeleton enhanced gan-based model for brush handwriting font
  generation.
\newblock {\em arXiv}, Apr 2022.
\newblock arXiv:2204.10484.

\bibitem{fogel2020}
S.~Fogel, H.~Averbuch-Elor, S.~Cohen, S.~Mazor, and R.~Litman.
\newblock Scrabblegan: Semi-supervised varying length handwritten text
  generation.
\newblock {\em arXiv}, Mar 2020.
\newblock arXiv:2003.10557.

\bibitem{wang2021}
C.~Wang, Y.~Tang, Z.~Jiang, and W.~Zhang.
\newblock An end-end method for handwritten xibo font generation.
\newblock In {\em 2021 4th International Conference on Artificial Intelligence
  and Pattern Recognition}, pages 662--668, Xiamen, China, Sep 2021. ACM.

\bibitem{elaraby2022}
N.~Elaraby, S.~Barakat, and A.~Rezk.
\newblock A conditional gan-based approach for enhancing transfer learning
  performance in few-shot hcr tasks.
\newblock {\em Sci. Rep.}, 12(1):16271, Sep 2022.

\bibitem{patil2022}
V.~Patil, T.~Ghosh, S.~Abdelhak, and C.-H.~S. Kuo.
\newblock Natural handwriting style generation with author adaptation.
\newblock In {\em 2022 IEEE International Conference on Systems, Man, and
  Cybernetics (SMC)}, pages 955--961, Prague, Czech Republic, Oct 2022. IEEE.

\bibitem{riaz2022}
N.~Riaz, H.~Arbab, A.~Maqsood, A.~Ul-Hasan, and F.~Shafait.
\newblock Conv-transformer architecture for unconstrained off-line urdu
  handwriting recognition.
\newblock 2022.

\bibitem{kotani2020}
A.~Kotani, S.~Tellex, and J.~Tompkin.
\newblock Generating handwriting via decoupled style descriptors.
\newblock {\em arXiv}, Sep 2020.
\newblock arXiv:2008.11354.

\bibitem{desousa2024}
A.~F. De~Sousa~Neto, B.~L.~D. Bezerra, G.~C.~D. De~Moura, and A.~H. Toselli.
\newblock Data augmentation for offline handwritten text recognition: A
  systematic literature review.
\newblock {\em SN Comput. Sci.}, 5(2):258, Feb 2024.

\bibitem{shonenkov2021}
A.~Shonenkov, D.~Karachev, M.~Novopoltsev, M.~Potanin, and D.~Dimitrov.
\newblock Stackmix and blot augmentations for handwritten text recognition.
\newblock {\em arXiv}, Aug 2021.
\newblock arXiv:2108.11667.

\bibitem{sareen2024}
B.~Sareen, R.~Ahuja, and A.~Singh.
\newblock Cnn-based data augmentation for handwritten gurumukhi text
  recognition.
\newblock {\em Multimed. Tools Appl.}, 83(28):71035--71053, Feb 2024.

\bibitem{zdenek2021}
J.~Zdenek and H.~Nakayama.
\newblock Jokergan: Memory-efficient model for handwritten text generation with
  text line awareness.
\newblock In {\em Proceedings of the 29th ACM International Conference on
  Multimedia}, pages 5655--5663, Virtual Event, China, Oct 2021. ACM.

\bibitem{shao2020}
Y.~Shao and C.-L. Liu.
\newblock Teaching machines to write like humans using l-attributed grammar.
\newblock {\em Eng. Appl. Artif. Intell.}, 90:103489, Apr 2020.

\bibitem{mustapha2022}
I.~B. Mustapha, S.~Hasan, H.~Nabus, and S.~M. Shamsuddin.
\newblock Conditional deep convolutional generative adversarial networks for
  isolated handwritten arabic character generation.
\newblock {\em Arab. J. Sci. Eng.}, 47(2):1309--1320, Feb 2022.

\bibitem{alonso2019}
E.~Alonso, B.~Moysset, and R.~Messina.
\newblock Adversarial generation of handwritten text images conditioned on
  sequences.
\newblock {\em arXiv}, Mar 2019.
\newblock arXiv:1903.00277.

\bibitem{luo2022}
C.~Luo, Y.~Zhu, L.~Jin, Z.~Li, and D.~Peng.
\newblock Slogan: Handwriting style synthesis for arbitrary-length and
  out-of-vocabulary text.
\newblock {\em arXiv}, Feb 2022.
\newblock arXiv:2202.11456.

\bibitem{elarian2015}
Y.~Elarian, I.~Ahmad, S.~Awaida, W.~G. Al-Khatib, and A.~Zidouri.
\newblock An arabic handwriting synthesis system.
\newblock {\em Pattern Recognit.}, 48(3):849--861, Mar 2015.

\bibitem{pippi2023}
V.~Pippi, S.~Cascianelli, and R.~Cucchiara.
\newblock Handwritten text generation from visual archetypes.
\newblock {\em arXiv}, Mar 2023.
\newblock arXiv:2303.15269.

\bibitem{kang2020icfhr}
L.~Kang, P.~Riba, M.~Rusinol, A.~Fornes, and M.~Villegas.
\newblock Distilling content from style for handwritten word recognition.
\newblock In {\em 2020 17th International Conference on Frontiers in
  Handwriting Recognition (ICFHR)}, pages 139--144, Dortmund, Germany, Sep
  2020. IEEE.

\bibitem{shonenkov2021strikethrough}
A.~Shonenkov, D.~Karachev, M.~Novopoltsev, M.~Potanin, D.~Dimitrov, and
  A.~Chertok.
\newblock Handwritten text generation and strikethrough characters
  augmentation.
\newblock {\em arXiv}, Dec 2021.
\newblock arXiv:2112.07395.

\bibitem{mattick2021}
A.~Mattick, M.~Mayr, M.~Seuret, A.~Maier, and V.~Christlein.
\newblock Smartpatch: Improving handwritten word imitation with patch
  discriminators.
\newblock volume 12821, pages 268--283, 2021.

\bibitem{kang2022}
L.~Kang, P.~Riba, M.~Rusiñol, A.~Fornés, and M.~Villegas.
\newblock Content and style aware generation of text-line images for
  handwriting recognition.
\newblock {\em IEEE Trans. Pattern Anal. Mach. Intell.}, 44(12):8846--8860, Dec
  2022.

\bibitem{davis2020}
B.~Davis, C.~Tensmeyer, B.~Price, C.~Wigington, B.~Morse, and R.~Jain.
\newblock Text and style conditioned gan for generation of offline handwriting
  lines.
\newblock {\em arXiv}, Sep 2020.
\newblock arXiv:2009.00678.

\bibitem{liu2021}
X.~Liu, G.~Meng, S.~Xiang, and C.~Pan.
\newblock Handwritten text generation via disentangled representations.
\newblock {\em IEEE Signal Process. Lett.}, 28:1838--1842, 2021.

\bibitem{lian2016}
Z.~Lian, B.~Zhao, and J.~Xiao.
\newblock Automatic generation of large-scale handwriting fonts via style
  learning.
\newblock In {\em SIGGRAPH ASIA 2016 Technical Briefs}, pages 1--4, Macau, Nov
  2016. ACM.

\bibitem{huu2021}
M.-K.~N. Huu, S.-T. Ho, V.-T. Nguyen, and T.~D. Ngo.
\newblock Multilingual-gan: A multilingual gan-based approach for handwritten
  generation.
\newblock In {\em 2021 International Conference on Multimedia Analysis and
  Pattern Recognition (MAPR)}, pages 1--6, Hanoi, Vietnam, Oct 2021. IEEE.

\bibitem{ding2023}
H.~Ding, B.~Luan, D.~Gui, K.~Chen, and Q.~Huo.
\newblock Improving handwritten ocr with training samples generated by glyph
  conditional denoising diffusion probabilistic model.
\newblock {\em arXiv}, May 2023.
\newblock arXiv:2305.19543.

\bibitem{haines2016}
T.~S.~F. Haines, O.~Mac~Aodha, and G.~J. Brostow.
\newblock My text in your handwriting.
\newblock {\em ACM Trans. Graph.}, 35(3):1--18, Jun 2016.

\end{thebibliography}

\end{document}